\documentclass{article}
\usepackage{amssymb}

\usepackage{arxiv}

\usepackage[utf8]{inputenc} 
\usepackage[T1]{fontenc}    
\usepackage{url}            
\usepackage{booktabs}       
\usepackage{amsfonts}       
\usepackage{nicefrac}       
\usepackage{microtype}      
\usepackage{lipsum}
\usepackage{graphicx}
\usepackage[colorlinks]{hyperref}
\usepackage{tabularx}
\usepackage{multirow}
\usepackage{amsmath} 
\usepackage{comment}
\usepackage{algorithm}
\usepackage{amsmath}
\usepackage{algpseudocode}
\usepackage{booktabs}
\usepackage{caption}
\usepackage{xcolor}
\usepackage{tcolorbox}
\tcbuselibrary{breakable} 
\usepackage{appendix}
\usepackage{subcaption}

\graphicspath{ {./images/} }
\newcommand{\argmin}{\operatornamewithlimits{arg\,min}}
\title{\textbf{SkyReels-V2}: Infinite-length Film Generative model}

\author{
 SkyReels Team \\
  Skywork AI \\
}

\begin{document}
\maketitle

\begin{abstract}
Recent advances in video generation have been driven by diffusion models and autoregressive frameworks, yet critical challenges persist in harmonizing prompt adherence, visual quality, motion dynamics, and duration: compromises in motion dynamics to enhance temporal visual quality, constrained video duration (5-10 seconds) to prioritize resolution, and inadequate shot-aware generation stemming from general-purpose MLLMs' inability to interpret cinematic grammar, such as shot composition, actor expressions, and camera motions. These intertwined limitations hinder realistic long-form synthesis and professional film-style generation. To address these limitations, we propose \textbf{SkyReels-V2}, an Infinite-length Film Generative Model, that synergizes Multi-modal Large Language Model (MLLM), Multi-stage Pretraining, Reinforcement Learning, and Diffusion Forcing Framework. Firstly, we design a comprehensive structural representation of video that combines the general descriptions by the Multi-modal LLM and the detailed shot language by sub-expert models. Aided with human annotation, we then train a unified Video Captioner, named \textbf{SkyCaptioner-V1}, to efficiently label the video data. Secondly, we establish progressive-resolution pretraining for the fundamental video generation, followed by a four-stage post-training enhancement: Initial concept-balanced Supervised Fine-Tuning (SFT) improves baseline quality; Motion-specific Reinforcement Learning (RL) training with human-annotated and synthetic distortion data addresses dynamic artifacts; Our diffusion forcing framework with non-decreasing noise schedules enables long-video synthesis in an efficient search space; Final high-quality SFT refines visual fidelity. Experiments demonstrate state-of-the-art performance in prompt adherence (especially the shot language), motion quality with sufficient dynamics, and film-style long-video generation capability, enabling several applications such as story generation, image-to-video synthesis, camera director and elements-to-video generation. All the code and models are available at \url{https://github.com/SkyworkAI/SkyReels-V2}.
\end{abstract}

\begin{figure}[t]
\centering
\includegraphics[width=\linewidth]{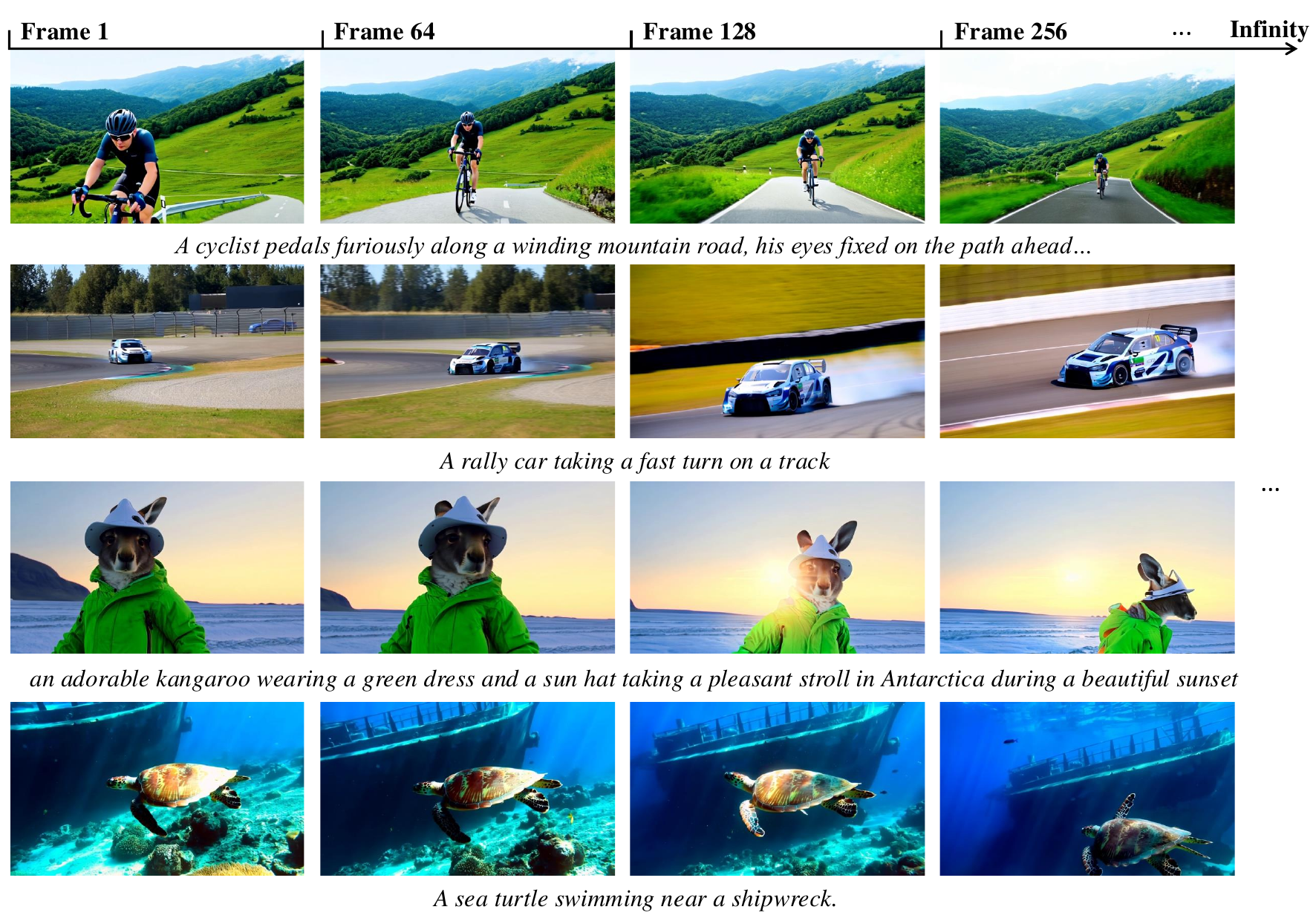}
\caption{\textbf{SkyReels-V2} produces stunningly realistic and cinematic high-resolution videos of virtually unlimited length. The model excels at maintaining visual consistency of the main subject across all frames, ensuring no distortion and delivering exceptional quality throughout the extended video sequences.}
\label{fig:demo_fig1}
\end{figure}

\section{Introduction}
Video generation has emerged as a pivotal domain in generative AI, enabling applications from creative content production to virtual simulation. While closed-source diffusion models have successful commercial applications, such as Sora~\cite{openai2024video}, Kling1.6~\cite{kuaishou2024kling}, Hailuo~\cite{minimax2024hailuo}, and Veo2~\cite{deepmind2024veo2}, open-source models struggles to reduce the performance gap with the closed-source models, among which Wan2.1~\cite{wan2025} shows a great improvement on the public benchmark and ranks, securing the No.1 position as of 2025-02-24 in V-bench1.0~\cite{huang2023vbench}. However, the commercial application of these video generative models 
for film makers still faces great challenges, including a minor text alignment in shot-language prompt following, lack of high-quality motion dynamics, and limited duration (typically 5s-10s). Several factors account for these limitations. Firstly, most of existing methods leverage the general MLLM to caption the video data which has a minor text alignment when handling the movie or film scenario with detailed shot language, such that the generation results will lose the professional movie performance. 
Secondly, optimization objectives in these models remain underexplored, leading to a poor motion quality that is important for the film maker. Standard denoising losses prioritize the frame-wise appearance learning and struggles with the temporal coherence as analysed in \cite{videoJam}. Though recent methods try the preference alignment methods to improve all metrics simutanously, such as semantics, aesthetics and motion dynamics. The weighting for each metric will be ill-defined, leading to a sub-optimal result. Furthermore, two primary methods dominates the video generation framework, including the diffusion models and autoregressive (AR) models. While diffusion models have set new benchmarks for visual quality through iterative denoising, and autoregressive (AR) models excel in temporal coherence, existing approaches struggle to harmonize these strengths. For instance, pure diffusion models often produce visually stunning but temporally fragmented outputs, while AR models suffer from error accumulation and degraded resolution. Due to these limitations, both of the methods cannot produce a long-duration video. To combine the merits of both high-fidelity diffusion methods and casual auto-regressive methods, some researchers proposes the diffusion-forcing transformers (DFoT) to bridge this gap but face critical limitations: DFoT’s combinatorial noise schedules lead to unstable training. 

To overcome these limitations, we propose the \textbf{SkyReels-V2} that synergizes Multi-modal Large Language model (MLLM), Multi-stage Pretraining, Reinforcement Learning, and Diffusion-forcing framework. Our approach begins with a careful ensemble design for video captioning. We propose a structural representations for a training video clip, which includes the subject type, subject appearance, expression, action, position, etc.  Some of the fields can be well understood by the general MLLM models like Qwen2.5-VL~\cite{Qwen2.5-VL}. These fields requires expert models, such as the shot type, shot angle, shot position, expression, and camera motions. To enhance the understanding of these fields, we train several expert models to achieve accurate descriptions. For efficiently labeling, we then distill the knowledge of both the general captioner and expert models into a unified MLLM model - \textbf{SkyCaptioner-V1}. The final text prompt is refined by a LLM to form the diverse description of video captions following the same original structural information. With these careful text description, we then pretrain a base diffusion model in progressive resolutions. After that we perform a first high-quality SFT stage with concept-balanced data to set up a good initialization for further optimizations. Then, inspired by the success application of Reinforcement Learning in LLM reasoning models like GPT-o1~\cite{gpt-o1} and deepseek-R1~\cite{deepseekR1}, we enhance the motion quality of the pretrained model through preference optimizations. To tackle the high cost of data annotation in RL, we propose a semi-auto pipeline to produce preference pairs. Besides, to unlock long-video synthesis and reduce convergence uncertainty, instead of pretraining the diffusion-forcing model from scratch, we propose a diffusion-forcing post-training where the pretrained full diffusion model will be finetuned into a diffusion-forcing model. To reduce the search space of denoising schedule as in \cite{sun2025ar},we use the non-decreasing noise schedule in the consecutive frames, which significantly reduce the Composition Space Size from $O(1e48)$ to $O(1e32)$. Finally, we train the model in the higher resolution and apply distill techniques to enable a high-quality commercial application. The proposed model also enables several applications including story generation, image-to-video synthesis, camera director and elements-to-video generation.

Extensive experiments demonstrate the superior performance of \textbf{SkyReels-V2} compared to current state-of-the-art methods. To the best of our knowledge, it represents the first open-source video generative model employing diffusion-forcing architecture that achieves the highest V-Bench score among publicly available models. Notably, our solution unlocks unprecedented infinite-length generation capabilities as depicted in Figure~\ref{fig:demo_fig1}. Human assessments conducted through the SkyReels-Bench benchmark further reveal that our model outperforms several closed-source alternatives and demonstrates comparable results with the leading video generative model in the field.
Our main contributions are summarized as follow:
\begin{itemize}
    \item Comprehensive video captioner that understand the shot language while capturing the general description of the video, which dramatically improve the prompt adherence. 
    \item Motion-specific preference optimization enhances motion dynamics with a semi-automatic data collection pipeline.
    \item Effective Diffusion-forcing adaptation enables the generation of ultra-long videos and story generation capabilities, providing a robust framework for extending temporal coherence and narrative depth.
    \item \textbf{SkyCaptioner-V1} and \textbf{SkyReels-V2} series models including diffusion-forcing, text2video, image2video, camera director and elements2video models with various sizes (1.3B, 5B, 14B) are open-sourced.
\end{itemize}

\begin{figure}[t]
\centering
\includegraphics[width=\linewidth]{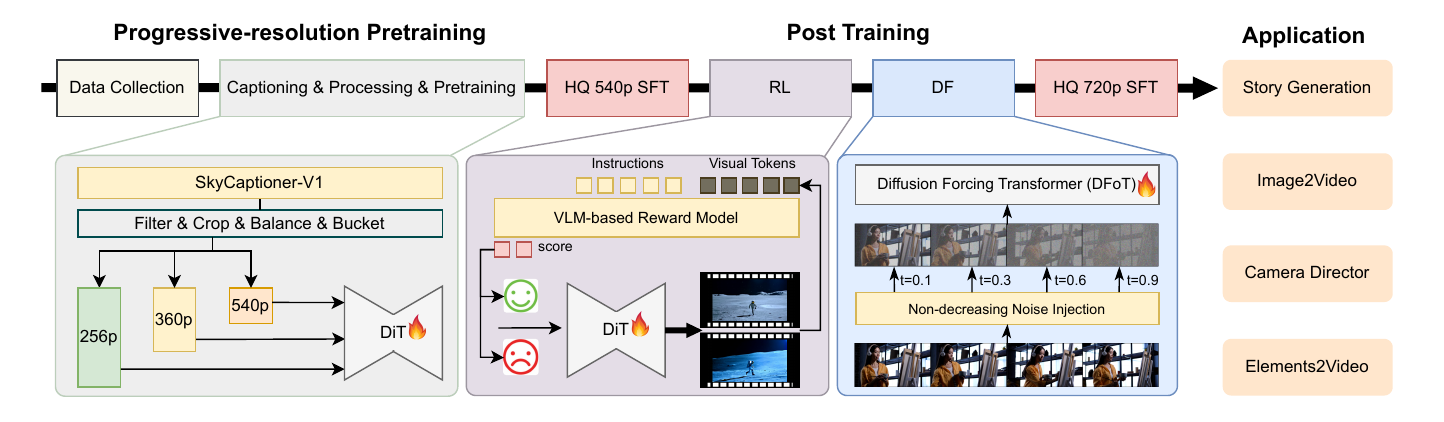}
\caption{Overview of the proposed method.}
\label{fig:overview}
\end{figure}

\section{Related Work}
\label{sec:related_work}
\subsection{Video Generative Models}
The field of video generation has witnessed remarkable advancements over the past year. While closed-source models like OpenAI Sora~\cite{openai2024video}, KuaiShou Kling~\cite{kuaishou2024kling}, MiniMax Hailuo~\cite{minimax2024hailuo}, RunwayML Gen-4~\cite{runwayml2024gen4}, and Google Veo2~\cite{deepmind2024veo2} have achieved commercial success, open-source alternatives are rapidly closing the performance gap. Early architectures predominantly employed 2D Spatial + 1D Temporal frameworks such as Make-A-Video~\cite{singer2022make}, AnimateDiff~\cite{guo2023animatediff}, Stable Video Diffusion~\cite{blattmann2023stable}, etc., which have gradually evolved into sophisticated 3D full-attention systems exemplified by Video Diffusion Models~\cite{ho2022video} and CogVideoX~\cite{yang2024cogvideox}. Recent open-source implementations including HunyuanVideo~\cite{kong2024hunyuanvideo}, StepVideo~\cite{ma2025stepvideot2vtechnicalreportpractice}, SkyReels-V1\cite{skyreels2025}, OpenSora-2.0~\cite{peng2025open}, and Wan2.1~\cite{wan2025} demonstrate progressively diminishing quality disparities with their proprietary counterparts.

These improvements stem from multi-faceted innovations: architectural transitions from U-Net~\cite{ronneberger2015u} to DiT~\cite{Peebles2022DiT} or MMDiT~\cite{esser2024scaling} structures, enhanced VAE implementations~\cite{esser2021taming,opensora,li2024wf,zhao2024cvvae,agarwal2025cosmos,kong2024hunyuanvideo,wan2025}, upgraded text encoders~\cite{radford2021learning,raffel2020exploring,kong2024hunyuanvideo,chung2023unimax}, and paradigm shifts from DDPM~\cite{ho2020denoising,Rombach_2022_CVPR} to flow matching~\cite{lipman2022flow,esser2024scaling} optimization. Concurrently, refined data processing pipelines and advancements in video captioning capabilities (GPT-4o~\cite{gpt-4o}, Qwen2.5-VL~\cite{Qwen2.5-VL}, Gemini 2.5~\cite{GeminiPro}, Tarsier2~\cite{yuan2025tarsier2advancinglargevisionlanguage}, etc.) have significantly contributed to quality enhancements. The frontier of research now extends to novel integrations of reinforcement learning, hybrid autoregressive-diffusion approaches, and long-form video generation techniques. These emerging directions promise to bridge the remaining gap towards achieving cinematic-quality video synthesis.
\subsection{Alignment on Diffusion models}
The success of Reinforcement Learning from Human Feedback (RLHF) in aligning large language models with human preferences~\cite{ouyang2022traininglanguagemodelsfollow} has inspired its adaptation to visual generation tasks. 
There are two main representative optimization algorithms: (1) Reward-Weighted Regression (RWR) methods~\cite{2021Advantage,0Aligning} employ reinforcement learning to optimize policy models by weighting trajectories with explicit reward models; (2) Direct Preference Optimization (DPO) strategies~\cite{2023Diffusion,2023Using,2024Step,liu2024videodpoomnipreferencealignmentvideo,ma2025stepvideot2vtechnicalreportpractice,liu2025improving} bypasses explicit reward modeling by directly optimizing preference data. These methods have been proved to successfully
enhance performances of diffusion models with human preferences, improving aesthetics and semantic consistency. 
Following the framework in~\cite{liu2025improving}, we apply DPO method on flow matching for incorporating human feedback. Unlike the previous works, we mainly focus on the motion quality, ignoring the text-alignment and the visual quality optimizations that will be improved by the other training stages.

Reward models play a important role in aligning generative models, as they are used to collect preference data. Early works employed metrics like CLIP scores~\cite{radford2021learning} and image quality scores~\cite{2021MUSIQ} as reward model to improve the assessment of visual quality and text alignment. Recent methods~\cite{liu2025improving,wang2025unifiedrewardmodelmultimodal,tong2025mjvideofinegrainedbenchmarkingrewarding} have started training Reward Models with human-annotated datasets, resulting in more accurate and direct outcomes. However, the generated data used in these methods are relatively outdated, leading to poor human alignment in motion quality of the trained Reward Models. Considering the high cost in collecting and annotating motion quality data, we propose a semi-automatic data collection pipeline to scale the motion-quality data, achieving significant improvements in alignment performance.

\subsection{Diffusion forcing framework}
While existing diffusion models have demonstrated remarkable success in video generation, they remain constrained to producing fixed-length sequences, lacking the unlimited sequence extension capability inherent in Large Language Models (LLMs) through autoregressive token prediction. Previous autoregressive approaches attempting to model video as next-token prediction suffer from error accumulation issues, leading to suboptimal performance compared to diffusion-based approaches. The emerging paradigm of Diffusion Forcing~\cite{chen2024diffusionforcingnexttokenprediction} addresses this by establishing a next-token prediction mechanism through independent noise level to form partial masking, thereby combining the high-quality generation of diffusion models with the infinite extension potential of autoregressive methods. However, the expanded search space in this framework introduces significant training challenges. To overcome this, AR-Diffusion~\cite{sun2025ardiffusionasynchronousvideogeneration} introduces a novel non-decreasing timestep constraint that systematically reduces the search space and stabilizes the training process. Furthermore, the History-Guided Video Diffusion~\cite{song2025historyguidedvideodiffusion} enhances temporal coherence by extending Classifier-Free Guidance (CFG)~\cite{ho2022classifierfreediffusionguidance} to accommodate variable-length context frame conditioning, significantly improving historical information utilization. CausVid~\cite{yin2025slowbidirectionalfastautoregressive} proposes an efficient adaptation strategy through DMD distillation~\cite{yin2024onestepdiffusiondistributionmatching,yin2024improveddistributionmatchingdistillation}, enabling direct conversion of pretrained bidirectional diffusion transformers into autoregressive diffusion forcing architectures without full retraining. Additionally, the Long Context Tuning framework~\cite{guo2025longcontexttuningvideo} implements a extensive approach: applying diffusion forcing at the scene level while maintaining full-sequence diffusion at the shot level, thereby enabling infinitely extensible story generation while preserving local visual quality. These innovations collectively advance the frontier of long-form video synthesis through synergistic integration of diffusion and auto-regressive paradigms.

\section{Methods}
\label{sec:headings}
In this section, we present a comprehensive overview of our methodology. Figure~\ref{fig:overview} illustrates the training framework. We begin by detailing the data processing pipeline in Sec.~\ref{sec:data_processing}, followed by an explanation of the Video Captioner architecture in Sec.~\ref{sec:video_captioner}. Next, we describe our multi-task pretraining strategy in Sec.~\ref{subsec:pretraining}. Subsequently, we elaborate on post-training optimization techniques in Sec.~\ref{subsec:posttraining}, including Reinforcement Learning in Sec.~\ref{subsec:RL}, Diffusion Forcing Training in Sec.~\ref{subsec:diffusion_forcing}, and High-quality Supervised Fine-Tuning (SFT) stages in Sec.~\ref{subsec:SFT}. We further outline the computational infrastructure for training and inference in Sec.~\ref{subsec:infrastructure}. To validate our approach, we conduct a systematic comparison with state-of-the-art baselines in Sec.~\ref{subsec:performance}. Finally, we demonstrate practical applications of the proposed model in Sec.~\ref{subsec:application}, including Story Generation, Image-to-Video Synthesis, Camera Director and Elements-to-Video Generation.
\subsection{Data Processing}
\label{sec:data_processing}

\begin{figure}[t]
\centering
\includegraphics[width=\linewidth]{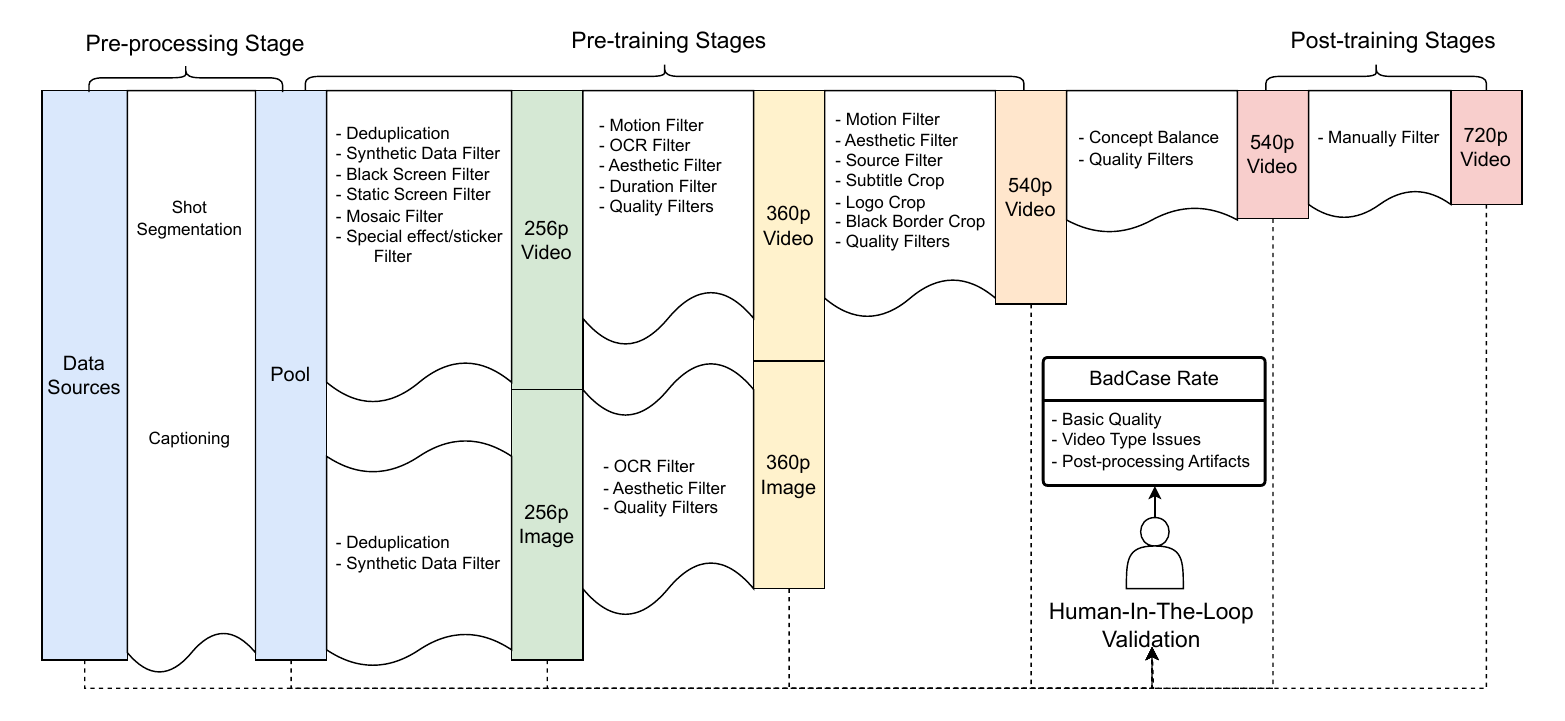}
\caption{Data Processing Pipeline.}
\label{fig:pipeline}
\end{figure}

Data processing stands as the cornerstone of video model training, and our framework integrates three critical components—\emph{Data Sources}, \emph{Processing Pipeline}, and \emph{Human-In-The-Loop Validation}—to ensure robust quality control. The Processing Pipeline, as depicted in Figure~\ref{fig:pipeline}, employs a progressive filtering strategy that transitions from loose to strict criteria, systematically reducing data volume while enhancing quality throughout the training process. This pipeline begins with raw inputs from diverse Data Sources, which are then processed through an automated pipeline designed to control the quality of samples by different filtering thresholds.
A key pillar of our pipeline is the integration of Human-In-The-Loop Validation, which focuses on manual evaluation of sampled data from both raw data sources and training samples in different stages. By conducting systematic sampling inspections at key stages—from initial data ingestion to pipeline outputs—this step ensures that ambiguous, erroneous, or non-compliant data are identified and rectified, ultimately safeguarding the final data quality critical for robust model training.

\subsubsection{Data Sources}
Given our objective to develop a film generative model, our multi-stage quality assurance framework integrates data from three primary sources: (1) The general-purpose dataset integrates open-source resources including Koala-36M~\cite{wang2024koala36mlargescalevideodataset}, HumanVid~\cite{humanvid}, along with additional web-crawled video resources from the internet.  (2) Self-collected media comprising 280,000+ films and 800,000+ TV episodes spanning 120+ countries (estimated total duration: 6.2M+ hours). (3) Artistic repositories featuring high-quality video assets from Internet. The raw dataset reaches $O(100M)$ scale, with different subsets utilized at various training stages based on quality requirements. We also collected $O(100M)$ concept-balanced image data to accelerate the establishment of generation capabilities during early training.
\subsubsection{Processing Pipeline}
\label{sub-sec:data-processing-pipeline}
As shown in Figure~\ref{fig:pipeline}, to obtain the training data pool, two pre-processing procedures are applied on the raw datas: Shot Segmentation and Captioning. After that, we handle the data quality issues using a series of data filters in different training stages. Through systematic analysis, we categorize data quality issues into three classes: {1) \emph{Basic quality}: low-resolution sources, low frame rates, black/white/static screens, camera shake, unstable motion, and arbitrary shot transitions. 2) \emph{Video type issues}: surveillance footage, game recordings, animation, meaningless content, and static videos. 3) \emph{Post-processing artifacts}: subtitles, logos, image editing, split screens, black/blurred borders, picture-in-picture, speed variations, and special effects/mosaics. The detailed definitions of these issues are shown in Table~\ref{tab:quality_issues}. Besides, we also use some data croppers to fix specific quality issues and perform the data balancing to ensure the generalization of the model. Pre-training Stages produce data for Multistage-pretraining in Sec.~\ref{subsec:pretraining}. Post-training Stages produce data for Post Training in Sec.~\ref{subsec:posttraining}.
\paragraph{Pre-processing Stage} The pre-processing stage consists of two processes: 1) \emph{Shot Segmentation}: All raw videos go through shot boundary detection using PyDetect and TransNet-V2~\cite{soucek2020transnetv2} and are splitted into single-shot video clips. 2) \emph{Captioning}: Segmented single-shot clips are annotated using our hierarchical captioning system as described in Section~\ref{sec:video_captioner}.  After the pre-processing stage, the training data pool will undergo a series of data filters, which set different thresholds for different training stages. Meanwhile, data croppers are introduced to fix several data quality issues.
\paragraph{Details of Data Filters} In this part, we explain the taxonomy and details of the data filters. 
Data Filters consist of Element Filters and Quality Filters to filter data for different training stages. Element Filters are used to justify the severity of specific quality issues. These filters are either classification-based filters to test the existence or classes of issues, or score-based filters that will set different thresholds for different quality requirements. \emph{Element Filters} include: 1) \emph{Black Screen 
Filter}: use heuristic rules to detect the data with black screen. 2) \emph{Static Screen Filter}: calculate a flow-based score to detect the static screen data. 3) \emph{Aesthetic Filter}: rely on the aesthetic model~\cite{aesthetic_predictor_v2_5} to get the a score. 
4) \emph{Deduplication}: To enhance the diversity of our pre-training set, we eliminate perceptually redundant clips from it by leveraging the similarity within the copy-detection embedding space~\cite{pizzi2022self}.
5) \emph{OCR Filter}: analyse the existence of text and calculate the occupancy ratio of the text and crop the data dependent on the training stage. 6) \emph{Mosaic Filter}: a trained expert model to detect the mosaic region. 7) \emph{Special effect/sticker Filter}: a trained expert model to identify the Special effect/sticker. Besides, we also include several \emph{Quality Filters}, such as the Video Quality Assessment (VQA) model~\cite{wu2022fastvqa, wu2023dover, end2endvideoqualitytool}, Image Quality Assessment (IQA) model~\cite{agnolucci2024arniqa}, and Video Training Suitability Score (VTSS)~\cite{wang2024koala36mlargescalevideodataset}. We will use these model after certain training stages and set different thresholds to filer data. Figure~\ref{fig:pipeline} illustrates the application of different data filters during different training stages.

\begin{figure}[htbp]
  \centering
  \includegraphics[width=0.8\textwidth]{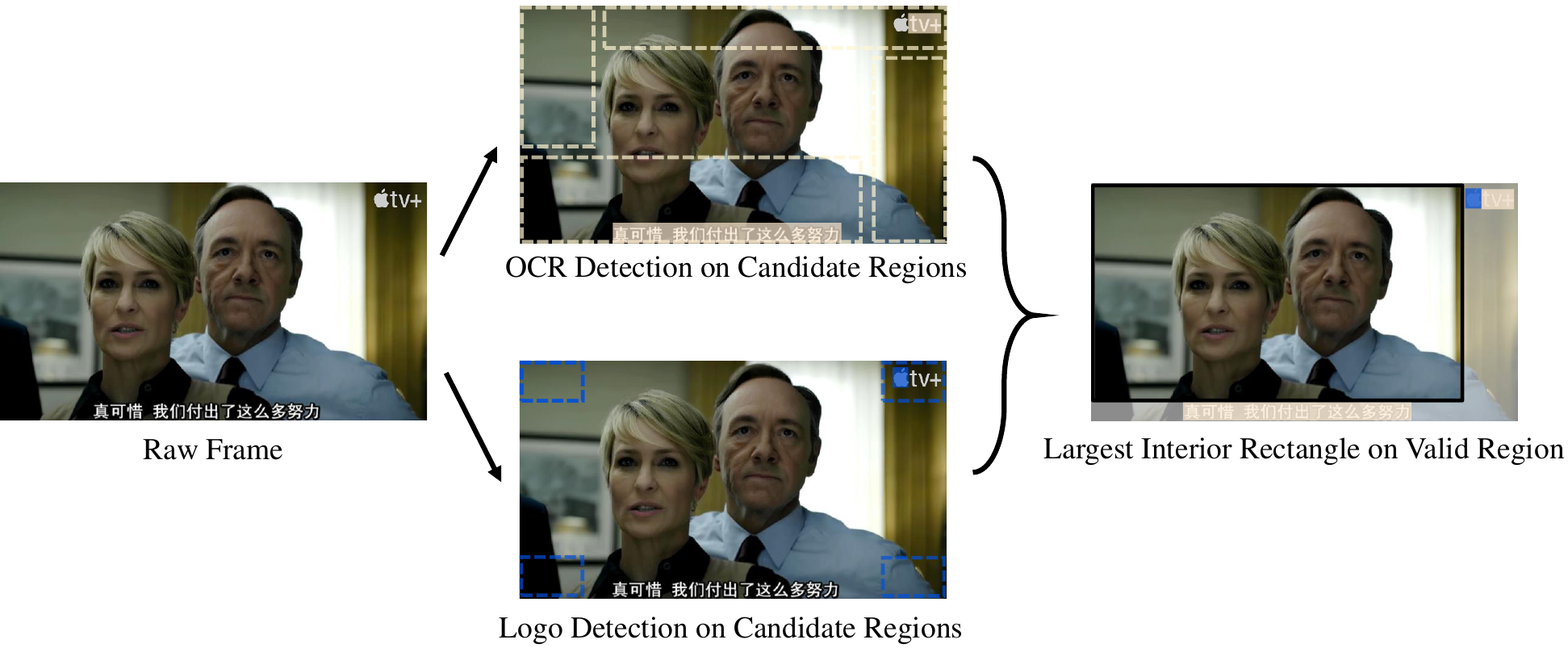}
  \caption{The pipeline of subtitle and logo cropping. Subtitles and logos are detected in candidate regions (dotted line). Then the largest interior rectangle is obtained beyond the the detection bounding boxes via Algorithm~\ref{alg:max_inner_rectangle}.}
  \label{fig:example}
\end{figure}

\paragraph{Details of Subtitle and Logo Crop}
Most of our training data comes from movies and TV series, which may have subtitles and channel logos that can affect the final video generation model's quality. Discarding this data directly is wasteful. To tackle this, we perform subtitle detection, logo detection, and video cropping sequentially on each video clip to remove overlays while keeping data quantity.
Before subtitle detection, we do Black Border Crop using heuristic - based methods to crop black borders, ensuring more reasonable subtitle position detection by providing cleaner data. 
For subtitle detection, we define four potential regions (top 20\%, bottom 40\%, left 20\%, and right 20\% of the frame borders) as candidate areas. Then the CRAFT model~\cite{CRAFT} is utilized to perform OCR detection on these regions in all video frames, and the coordinates of the OCR bounding boxes are recorded. Similarly, for logo detection, we focus on four corner regions (each covering 15\% of frame width/height) and employ the MiniCPM-o model~\cite{minicpm} to detect and record the logo coordinates.
In the video cropping phase, we first construct a binary matrix that matches the dimensions of the video frame, where the detected subtitle/logo regions are marked 0 and other areas 1. We then apply a monotonic stack-based algorithm (as detailed in Algorithm~\ref{alg:max_inner_rectangle}) to identify the largest interior rectangle containing only 1s. If such a rectangle covers over 80\% area of the original frame, and its aspect ratio is close to that of the original frame, then all frames will be cropped according to the coordinates of the rectangle and saved as a new video clip, while non-compliant data will be discarded. The entire video processing pipeline is illustrated in Fig.~\ref{fig:example}.

\paragraph{Data Balancing in Post Training}
\label{para:databalnce}
In post training stage, we begin to perform a detailed concept balancing using subject categories from the captioner, resulting in a 50\% data volume reduction. The comparison of the unbalanced and balanced concept grouped by main subject type is shown in Figure~\ref{fig:compare_unbalanced_balanced_data}. After concept balanced, we also calculated the distribution of each sub-type under each primary type. Table~\ref{tab:primary_subtype_ratios} provides a detailed breakdown of the sub-types statistics for the top five primary types.

\subsubsection{Human-In-The-Loop Validation}
Human-In-The-Loop Validation involves manual visual checks at every stage of data production—\emph{Data Sources}, \emph{Shot Segmentation}, \emph{Pre-training} and \emph{Post-training}—to ensure high-quality data for model training. For Data Sources, humans subjectively assess if the raw data is suitable for use. During Shot Segmentation, reviewers check samples to ensure less than 1\% of shots have errors like wrong transitions. In Pre-training, data is filtered and 0.01\% (1 in 10,000 samples) is manually checked to meet strict limits: overall bad cases (problems like poor quality, wrong content type, or processing errors) must be under 15\%, with subcategories like basic quality issues <3\%, video type issues <5\% and post-processing flaws <7\%. For Post-training, the same 0.1\% sample rate (1 in 1000 samples) applies but with tighter rules: total bad cases must be under 3\%, including basic quality <0.5\%, video type issues <1\%, and post-processing flaws <1.5\%. 
We determine the usability of Data Sources batches by leveraging the bad cases rates derived from manual checks. If the bad cases rate of a particular batch surpasses the predefined threshold, appropriate actions such as discarding or further refining the batch will be taken. Moreover, filter parameters are adjusted in accordance with the characteristics of diverse data sources. For instance, filters related to quality are tightened for data sources with a higher incidence of quality-related issues. This step-by-step manual evaluation at each stage ensures data quality stays high, helping the model train effectively.

\begin{table}[t]
\centering
\caption{Data Quality Issue Categories and Definitions}
\label{tab:quality_issues}
\begin{tabular}{lp{4cm}p{7cm}}
\toprule
\textbf{Category} & \textbf{Issue} & \textbf{Definition} \\
\midrule
\multirow{6}{*}{Basic Quality} 
& Low-resolution sources & Video sources with insufficient pixel density, typically below 720p resolution \\
& Low frame rates & Videos with frame rates below 16fps causing choppy motion \\
& Black/white/static screens & Frames containing blank screens or frozen images \\
& Camera shake & Unintentional camera movement causing unstable footage \\
& Unstable motion & Irregular object/camera movement creating visual discomfort \\
& Arbitrary shot transitions & Abrupt or mismatched scene changes without logical continuity \\
\midrule

\multirow{5}{*}{Video Type Issues}
& Surveillance footage & CCTV-style recordings with fixed angles and timestamps \\
& Game recordings & Screen captures of video game gameplay \\
& Animation & Computer-generated or hand-drawn non-live-action content \\
& Meaningless content & Videos lacking coherent narrative or visual purpose \\
& Static videos & Footage with minimal motion (e.g., still images with audio) \\
\midrule

\multirow{8}{*}{Post-processing Artifacts}
& Subtitles & Text overlays added during editing \\
& Logos & Watermarks or channel identifiers superimposed on video \\
& Image editing & Color grading, filters, or digital alterations \\
& Split screens & Multiple video streams shown simultaneously \\
& Black/blurred borders & Non-content areas added during post-production \\
& Picture-in-picture & Secondary video inset within main footage \\
& Speed variations & Altered playback speed (slow/fast motion) \\
& Special effects/mosaics & Added visual elements or pixelation overlays \\
\bottomrule
\end{tabular}
\end{table}

\begin{figure}[t]
\centering
\includegraphics[width=\linewidth]{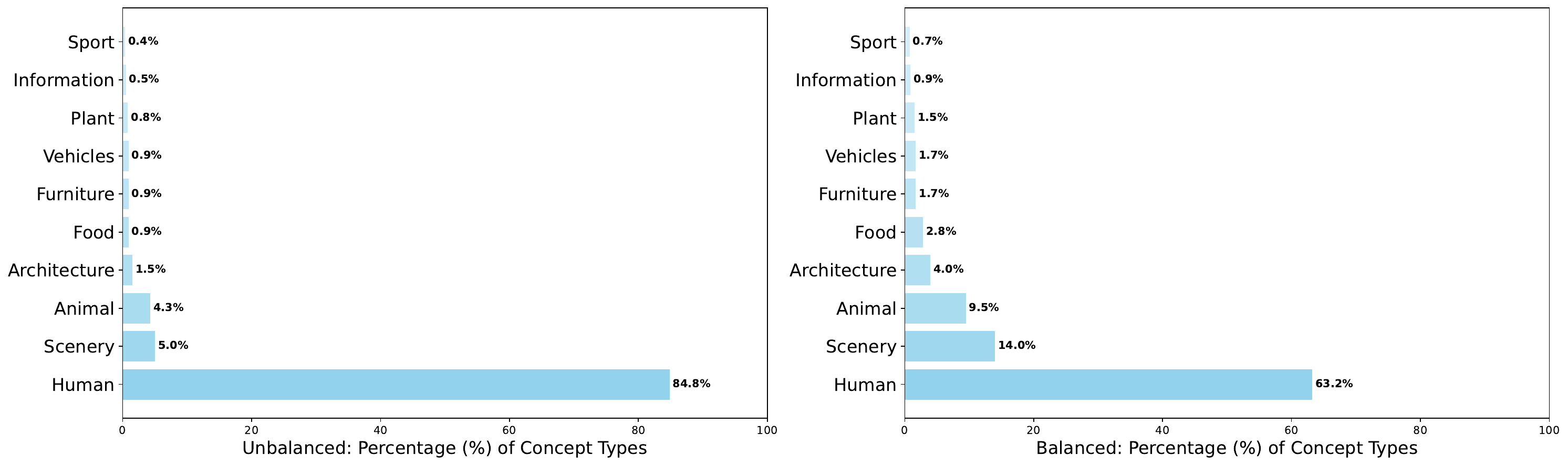}
\caption{Comparison of unbalanced (left) and balanced (right) concept distribution .}
\label{fig:compare_unbalanced_balanced_data}
\end{figure}

\begin{table}[htbp]
\centering
\addtolength{\tabcolsep}{11pt}
\caption{Statics of top5 primary types and its sub-types ratios}
\label{tab:primary_subtype_ratios}
\begin{tabular}{llllll}
\toprule
Primary type & Sub-type & Ratio & Primary type & Sub-type & Ratio \\
\midrule
\multirow{6}{*}{Human} 
& Man & 55.2\% 
& \multirow{6}{*}{Animal} 
& Mammal & 55.4\% \\ 
& Woman & 40.8\% 
& & Bird & 18.5\% \\ 
& Girl & 1.9\% 
& & Aquatic Life & 13.4\% \\ 
& Boy & 1.4\% 
& & Insect & 7.5\% \\ 
& Child & 0.5\% 
& & Reptile & 5.2\% \\ 
& Baby & 0.2\% & \\
\midrule
\multirow{15}{*}{Scenery} 
& Mountain & 17.9\% 
& \multirow{6}{*}{Architecture} 
& Historical & 41.3\% \\ 
& Seascape & 16.4\% 
& & Commercial & 19.1\% \\ 
& Urban & 12.0\% 
& & Industrial & 16.3\% \\ 
& River & 10.6\% 
& & Residential & 12.3\% \\ 
& Beach & 7.6\% 
& & Religious & 11.1\% \\ 
\cmidrule(lr){4-6} 
& Road & 6.1\% 
& \multirow{8}{*}{Food} 
& Snack & 35.8\% \\  
& Lake & 6.1\% 
& & Dessert & 19.4\% \\ 
& Sky & 6.1\% 
& & Fruit & 11.3\% \\ 
& Forest & 5.7\% 
& & Meat & 10.7\% \\ 
& Volcano & 4.1\% 
& & Vegetable & 8.6\% \\ 
& Desert & 2.6\% 
& & Seafood & 6.9\% \\ 
& Valley & 2.0\% 
& & Dairy & 4.7\% \\
& Canyons & 1.7\% 
& & Poultry & 3.7\% \\ 
& Cloud & 1.2\% 
& & \\ 
\bottomrule
\end{tabular}
\end{table}

\subsection{Video Captioner}
\label{sec:video_captioner}
\begin{figure}
    \centering
    \includegraphics[width=\linewidth]{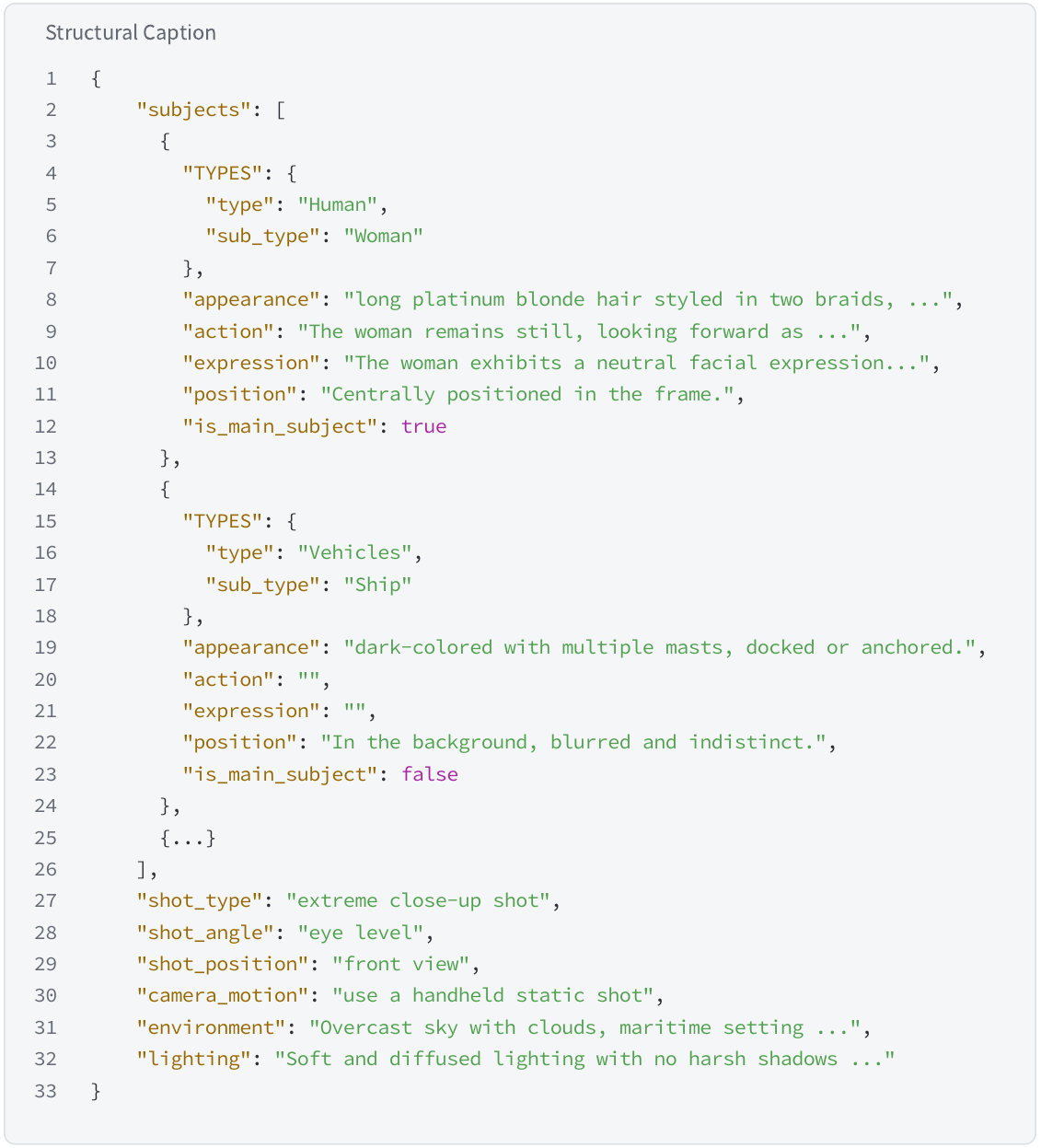}
    \caption{The design and demonstration of our structural caption}
    \label{fig:struct_caption}
\end{figure}

Our Video Captioner aims to generate precise video captions by integrating structured caption formats with specialized sub-expert captioners. Its objectives include: 1) Correcting errors or hallucinated information from Multi-modal Large Language model (MLLM). 2) Continuously optimizing dynamic video elements (e.g., shot information, expressions, and camera motions). 3) Dynamically adjusting caption length based on application scenarios (text-to-video or image-to-video). To achieve these objectives, we design a structural caption, as shown in Figure~\ref{fig:struct_caption}, which provide multi-dimensional descriptive information from various perspectives, including: 1) \emph{Subjects}: Main and secondary entities with attributes like appearance, action, expression, position, and hierarchical categories/types (e.g., Animal → Mammal). 2) \emph{Shot Metadata}: shot type, shot angle, shot position, camera motion, environment, lighting, etc. We use the base model Qwen2.5-VL-72B-Instruct to generate these initial structural information. However, some information will be replaced by the results of the expert captioners to get more precise descriptions. Finally, we generate the final captions by fusions of the structured data for different models. 1) \emph{Text-to-video}: Produces dense descriptions. 2) \emph{Image-to-video}: Focuses on "subject + temporal action/expression + camera motion.". Each fields of captions follow a 10\% drop rate to adapt different users situations as users might not give a precise description for each fields. The caption fusion details are displayed in Appendix~\ref{app:skyreels-system-prompt}. 

\subsubsection{Sub-expert Captioner}
\paragraph{Shot Captioner} The shot captioner consists of three sub-captioners that describe different aspects of a shot. It includes shot type, shot angle, and shot position. We define these aspects as classification problems. 1) \emph{Shot type}: close-up shot, extreme close-up shot, medium shot, long shot, and full shot. 2) \emph{Shot angle}: eye angle shot, high angle shot, low angle shot. 3) \emph{Shot position}: back view, front view, over the head, over the shoulder, point of view and side view.

Our training methodology employs a carefully designed two-phase approach to develop robust shot classifiers. In the initial phase, we train a preliminary classifier using web images to establish baseline performance (we use class labels as trigger word to crawl data from web). This low-precision model serves primarily to extract balanced real-world scene data across all target categories from our film datasets. The second phase focuses on developing high-precision expert classifiers through manual annotation of real film data, with each category containing 2,000 carefully labeled samples. These annotated samples form the training set for our final high-precision classifiers, which are specifically optimized for accurate shot type, shot angle and shot position classification in real film videos. This multi-stage training approach ensures both category balance in our training datasets and high classification accuracy for production applications.

To evaluate the performance of our three classifiers: shot type, shot angle, and shot position. We constructed a balanced test set containing 100 manually annotated samples per label. The evaluation results showed average accuracies of 82.2\% for shot type classification, 78.7\% for shot angle classification, and 93.1\% for shot position classification. While the shot position classifier achieved strong performance, the shot type and shot angle classifiers indicate potential for improvement in future, particularly through enhanced data balance and higher-quality annotations for scene and angle classification tasks.

\paragraph{Expression Captioner}
The expression captioner provides detailed descriptions of human facial expressions, focusing on several key dimensions: 1) \emph{Emotion Label}: Emotions are categorized into seven common types, \textit{i.e.}, \textit{neutral}, \textit{anger}, \textit{disgust}, \textit{fear}, \textit{happiness}, \textit{sadness}, and \textit{surprise}. 2) \emph{Intensity}: The strength of the emotion is quantified, such as "slight anger," "moderate joy," or "extremely surprised," indicating the emotion's magnitude. 3) \emph{Facial Features}: Physical characteristics contributing to emotional expression, including eye shape, eyebrow position, mouth curvature, wrinkles, and muscle movements. 4) \emph{Temporal Description}: Captures dynamic changes in emotional displays over time, focusing on how emotions evolve and the timing of these changes within a video.

The expression caption generation consists of two phases: 1) We first detect and crop human faces and classify their emotions using our emotion classifier. 2) We then input both the emotion labels and video frames into a VLM model to generate detailed expression captions.
Specifically, we adapt the framework of S2D~\cite{S2D} and train the model using \(\sim\)10k in-house datasets, focusing on both human and non-human characters. For the VLM model, we use InternVL2.5 to generate frame-wise descriptions with emotion labels as a prior and we employ a chain-of-thought prompting strategy to refine the descriptions and generate the final expression caption.


To validate the performance of our emotion classifier, we compiled a test set of 1,200 videos. The classifier achieved an average precision of 85\% across all emotion categories. For the expression captioner, we gathered 560 video samples and enlisted human annotators to evaluate the model's effectiveness across four key dimensions. The captioner demonstrated an accuracy of 88\% for emotion labeling, 95\% for emotion intensity assessment, 85\% for facial feature identification, and 93\% for temporal description accuracy. These results highlight the robustness of our model in capturing nuanced emotional and expressive details in video content.

\paragraph{Camera Motion Captioner} Our framework employs a hierarchical classification strategy for camera motions through a three-stage processing pipeline integrating Motion Complexity Filtering, Single-type Motion Modeling and Single-type Motion Data Curation. 1) \emph{Motion Complexity Filtering}: This stage eliminates trivial and overly complex motions through dual detection mechanisms. A binary static shot detector (95\% accuracy) first screens motionless clips, followed by specialized classifiers for irregular patterns (handheld jitter, subject tracking, abrupt shifts) trained on manually labeled data. Surviving clips proceed as standard single-type motions. 2) \emph{Single-type Motion Modeling}: We parameterize motions using 6DoF coordinates (translation x/y/z; rotation roll/pitch/yaw), each axis discretized into negative (-), neutral (0), or positive (+) states. Coupled with three speed tiers (slow: <5\%, medium: 5-20\%, fast: >20\% frame displacement/sec), this creates 2,187 distinct motion combinations. Training data combines manual annotations and synthetic samples. 3) \emph{Single-type Motion Data Curation}: We implement five-cycle active learning for efficient annotation scaling. Starting with O(10k) human-annotated samples for baseline training, we iteratively predict labels on 100k unlabeled data, balance-sample 10k predictions for verification, then refine models through fine-tuning. This process yields 93k high-confidence samples supplemented with 16k synthetic data balanced across motion axes. The synthetic data ensures equal positive/negative state representation for each DoF axis. All data trains classification-based captioners for motion recognition.

Evaluation on a 15k-video test set achieves prediction accuracies of: 89\% for single-type motions, 78\% (handheld), 83\% (subject-following), and 81\% (abrupt shifts) for complex motions, and 95\% accuracy for static shot detection.

\subsubsection{SkyCaptioner-V1: A Structured Video Captioning Model}
\textbf{SkyCaptioner-V1} serves as our final video captioning model for data annotation. This model is trained on the captioning result from the base model Qwen2.5-VL-72B-Instruct and the sub-expert captioners on a balanced video data. The balanced video data is a carefully curated dataset of approximately 2 million videos—selected from an initial pool of 10 million samples to ensure conceptual balance and annotation quality.

Built upon the Qwen2.5-VL-7B-Instruct model, \textbf{SkyCaptioner-V1} is fine-tuned to enhance performance in domain-specific video captioning tasks. To compare the performance with the SOTA models, we conducted a manual assessment of accuracy across different captioning fields using a test set of 1,000 samples. Table~\ref{tab:captioner_comparison} presents the detailed accuracy metrics of each field in the structural caption. The proposed \textbf{SkyCaptioner-V1} achieves the highest average accuracy among the baseline models, and show a dramatic result in the shot related fields.

\begin{table}[htbp]
    \centering
\caption{Comprehensive Model Performance Comparison on Visual Understanding test set (All the models use the same system prompt of generating structural caption for video in Appendix~\ref{app:skyreels-system-prompt}. For the \textit{Tarsier2-recap-7B} baseline, we implement a caption-to-json converter because it can not output structural format directly for their supervised fine-tuning method.)}
    \label{tab:captioner_comparison}
\begin{tabular}{l|c|c|c|c}
\hline
model & Qwen2.5-VL-7B-Ins.& Qwen2.5-VL-72B-Ins.& Tarsier2-recap-7B& \textbf{SkyCaptioner-V1} \\
\hline
Avg accuracy & 51.4\% & 58.7\% & 49.4\% & \textbf{76.3}\% \\
\hline
shot type & 76.8\% & 82.5\% & 60.2\% & \textbf{93.7}\% \\
shot angle & 60.0\% & 73.7\% & 52.4\% & \textbf{89.8}\% \\
shot position & 28.4\% & 32.7\% & 23.6\% & \textbf{83.1}\% \\
camera motion & 62.0\% & 61.2\% & 45.3\% & \textbf{85.3}\% \\
expression & 43.6\% & 51.5\% & 54.3\% & \textbf{68.8}\% \\
\hline
TYPES\_type & 43.5\% & 49.7\% & 47.6\% & \textbf{82.5}\% \\
TYPES\_sub\_type & 38.9\% & 44.9\% & 45.9\% & \textbf{75.4}\% \\
appearance & 40.9\% & 52.0\% & 45.6\% & \textbf{59.3}\% \\
action & 32.4\% & 52.0\% & \textbf{69.8}\% & 68.8\% \\
position & 35.4\% & 48.6\% & 45.5\% & \textbf{57.5}\% \\
is\_main\_subject & 58.5\% & 68.7\% & 69.7\% & \textbf{80.9}\% \\
environment & 70.4\% & \textbf{72.7}\% & 61.4\% & 70.5\% \\
lighting & 77.1\% & \textbf{80.0}\% & 21.2\% & 76.6\% \\
\hline
\end{tabular}
\end{table} 

\paragraph{Training details} We adopt Qwen2.5-VL-7B-Instruct as our base model and train it with a global batch size of 512 distributed across 64 NVIDIA A800 GPUs using 4 micro batch size and 2 gradient accumulation steps. The model is optimized using AdamW with a learning rate of 1e-5 and trained for 2 epochs, with the best-performing checkpoint selected based on comprehensive evaluation metrics from our test set. This training configuration ensures stable convergence while maintaining computational efficiency for large-scale video captioning tasks.

\subsection{Multistage-pretraining}
\label{subsec:pretraining}
We adopt the model architecture from Wan2.1\cite{wan2025} and only train the DiT from scratch while retaining the pretrained weight of other components including VAE and text encoder. Then, we also use the Flow Matching framework \cite{lipman2022flow,esser2024scaling} to train our video generation model. This approach transforms a complex data distribution into a simple Gaussian prior through continuous-time probability density paths, enabling efficient sampling via ordinary differential equations (ODEs). 
\paragraph{Training Objective:}
Given a latent representation $\mathbf{x}_1$ (image or video), we sample a timestep $t \in [0, 1]$ from a logit-normal distribution \cite{esser2024scaling}.  Then, initialize noise $\mathbf{x}_0 \sim \mathcal{N}(0, \mathbf{I})$. and construct the intermediate latent $\mathbf{x}_t$ via linear interpolation:  
   \begin{equation}
     \mathbf{x}_t = t \mathbf{x}_1 + (1 - t) \mathbf{x}_0.
   \end{equation}  
Compute the ground-truth velocity vector $\mathbf{v}_t$ as:  
   \begin{equation}
     \mathbf{v}_t = \frac{d\mathbf{x}_t}{dt} = \mathbf{x}_1 - \mathbf{x}_0.
   \end{equation}  

The model predicts the velocity field $\mathbf{u}_\theta(\mathbf{x}_t, \mathbf{c}, t)$, which guides the sample $\mathbf{x}_t$ towards the sample $\mathbf{x}_1$ and is conditioned on text embeddings $\mathbf{c}$ (e.g., 512-dim $\text{umT5}$ features), by minimizing the loss function \textit{L}:  
\begin{equation}
  \mathcal{L} = \mathbb{E}_{t, \mathbf{x}_0, \mathbf{x}_1, \mathbf{c}} \left[ \| \mathbf{u}_\theta(\mathbf{x}_t, \mathbf{c}, t) - \mathbf{v}_t \|^2 \right],
\end{equation}  
Following this training objective, we first design a dual-axis Bucketing framework and FPS normalization method to normalize all the data. After that, we perform three-stage pretrainings with a progressively increasing resolution. 
\paragraph{Dual-axis Bucketing framework and FPS normalization}
Following the data processing outlined in Section~\ref{sec:data_processing}, we address the spatiotemporal heterogeneity of video data through a dual-axis bucketing framework. This framework organizes training samples along two orthogonal dimensions: temporal duration bins ($B_T$ divisions) and spatial aspect ratio categories ($B_{AR}$ divisions), forming a $B_T$ × $B_{AR}$ matrix of mutually exclusive buckets. To optimize GPU memory utilization while preventing OOM failures, we implement adaptive batch sizing through empirical profiling - each bucket is assigned a distinct maximum batch capacity based on its duration and aspect ratio. During data preprocessing, samples are mapped to their nearest bucket. Throughout model training, distributed compute nodes employ stochastic bucket sampling to dynamically assemble mini-batches, ensuring continuous variation in input resolutions and temporal spans. Building upon the dual-axis bucketing system ($B_T$ temporal bins × $B_{AR}$ spatial aspect ratio categories), we extend the framework with temporal frequency adaptation. Videos undergo FPS normalization through a residue-aware downsampling protocol: For each sample, we compute modulus remainders relative to target frequencies (16/24 FPS), selecting the frequency with minimal remainder as the resampling basis. This mathematical formulation:
$f_{\mathrm{target}} = \argmin_{f \in \{16,24\}} \left( \mathrm{original\_fps} \bmod f \right)$
ensures optimal temporal alignment while preserving motion semantics. Resampled videos are subsequently bucketed using the established duration-aspect ratio matrix. To disentangle frame rate dependencies, we augment the DiT architecture with learnable frequency embeddings that interact additively with timestep embeddings. These learnable frequency embeddings will be abandoned after we use FPS-24 only video data in high-quality SFT stage.
\paragraph{Pretraining Stage1}
We first pretrain on low-resolution data (256p) to capture the basic generation ability. In this stage, we propose a joint image-video training and support different aspect ratio and frame length. We implement rigorous data filtering to remove low-quality and synthetic data, and perform deduplication to ensure data diversity. This low-resolution stage helps the model to learn the low-frequency concepts from a larger amount of samples. The model trained at this stage demonstrates fundamental video generation capabilities, though the generated videos remain relatively blurry.
\paragraph{Pretraining Stage2}
In this stage, we continue with joint image-video training but increase the resolution to 360p. We apply more sophisticated data filtering strategies, including duration filtering, motion filtering, OCR filtering, aesthetic filtering and quality filtering. After this training stage, the clarity of generated videos shows significant improvement.
\paragraph{Pretraining Stage3}
We further scale up the resolution to 540p in this final pretraining stage, focusing exclusively on video objectives. We implement more stringent motion, aesthetic and quality filtering criteria to ensure high-quality training data. Moreover, we introduce source filtering to remove user-generated content while preserving cinematic data. This approach enhances the visual quality of generated videos and significantly improves the model's capability to generate realistic human videos with superior texture and cinematic qualities. 
\paragraph{Pretraining Settings}
For optimization, we employ the AdamW optimizer throughout all pretraining stages. In Stage 1, we initialize the learning rate at 1e-4 with weight decay set to 0. Once the loss converges to a stable range, we adjust the learning rate to 5e-5 and introduce weight decay at 1e-4. In Stages 2 and 3, we further reduce the learning rate to 2e-5.
\subsection{Post Training}
\label{subsec:posttraining}
The post training is the key stage to improve the overall performance of the model. Our post training consists of four stages: high-quality SFT in 540p, Reinforcement Learning, Diffusion Forcing Training, and high-quality SFT in 720p. For efficiency, the first three post training are performed in 540p resolution, while the last stage is performed in 720p resolution. The high-quality SFT in 540p leverage the balanced data to improve the overall performance, setting a better initialization for the following stages. To enhance the motion quality, we will rely on the Reinforcement learning instead of standard diffusion loss. In this stage, we propose a semi-automatic pipeline to collect preference data from both human and models. Furthermore, we propose the diffusion forcing training stage, in which we transform the full-sequence diffusion model into a diffusion forcing model that applies frame-specific noise levels such that enables a various length video generation ability. After that, we have the high-quality SFT stage in 720p, which increase the generative resolution from 540p to 720p.
\subsubsection{Reinforcement Learning}
\label{subsec:RL}
Inspired by the previous success in LLM~\cite{gpt-o1,deepseekR1}, we propose to enhance the performance of the generative model by Reinforcement Learning. Specifically, we focus on the motion quality because we find that the main drawback of our generative model is: 1) the generative model does not handle well with large, deformable motions (Fig.~\ref{fig:vertical-compare}.a, Fig.~\ref{fig:vertical-compare}.b). 2) the generated videos may violate the physical law (Fig.~\ref{fig:vertical-compare}.c).

To avoid the degradation in other metrics, such as text alignment and video quality, we ensure the preference data pairs have comparable text alignment and video quality, while only the motion quality varies. This requirement poses greater challenges in obtaining preference annotations due to the inherently higher costs of human annotation. To address this challenge, we propose a semi-automatic pipeline that strategically combines automatically generated motion pairs and human annotation results. This hybrid approach not only enhances the data scale but also improves alignment with human preferences through curated quality control. Leveraging this enhanced dataset, we first train a specialized reward model to capture the generic motion quality differences between paired samples. This learned reward function subsequently guides the sample selection process for Direct Preference Optimization (DPO), enhancing the motion quality of the generative model.

\begin{figure}[htbp]
  \centering
  \begin{subfigure}[t]{\textwidth} 
    \centering
    \includegraphics[width=0.9\linewidth, height=8cm, keepaspectratio]{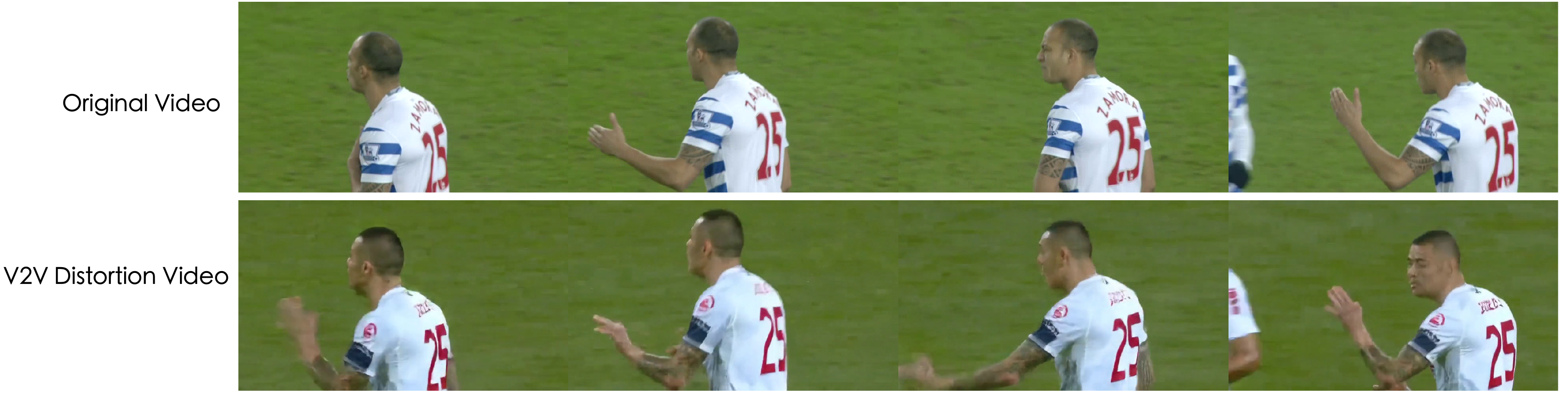}
    \captionsetup{justification=raggedright, singlelinecheck=false}
    \caption{An example of V2V distortion: The character's face undergoes slight corruption in the generated video.}
    \label{fig:vert1a}
  \end{subfigure}
  
  \vspace{0.5cm} 
  
  \begin{subfigure}[t]{\textwidth}
    \centering
    \includegraphics[width=0.9\linewidth, height=8cm, keepaspectratio]{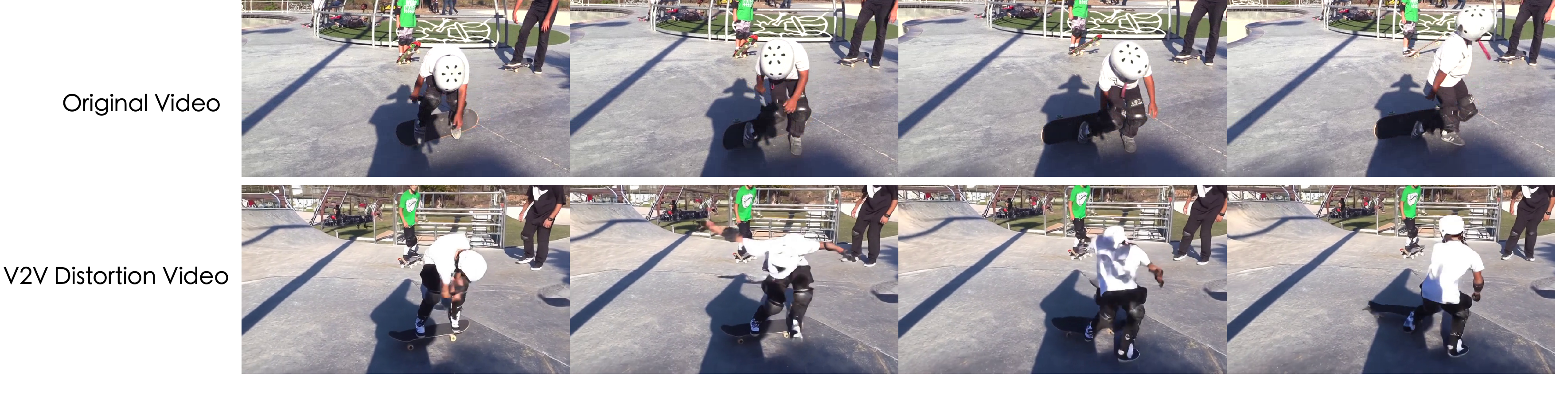}
    \captionsetup{justification=raggedright, singlelinecheck=false}
    \caption{An example of I2V distortion: The person in the generated video experiences severe body deformation.}
    \label{fig:vert1b}
  \end{subfigure}

  \vspace{0.5cm} 
  
  \begin{subfigure}[t]{\textwidth}
    \centering
    \includegraphics[width=0.9\linewidth, height=8cm, keepaspectratio]{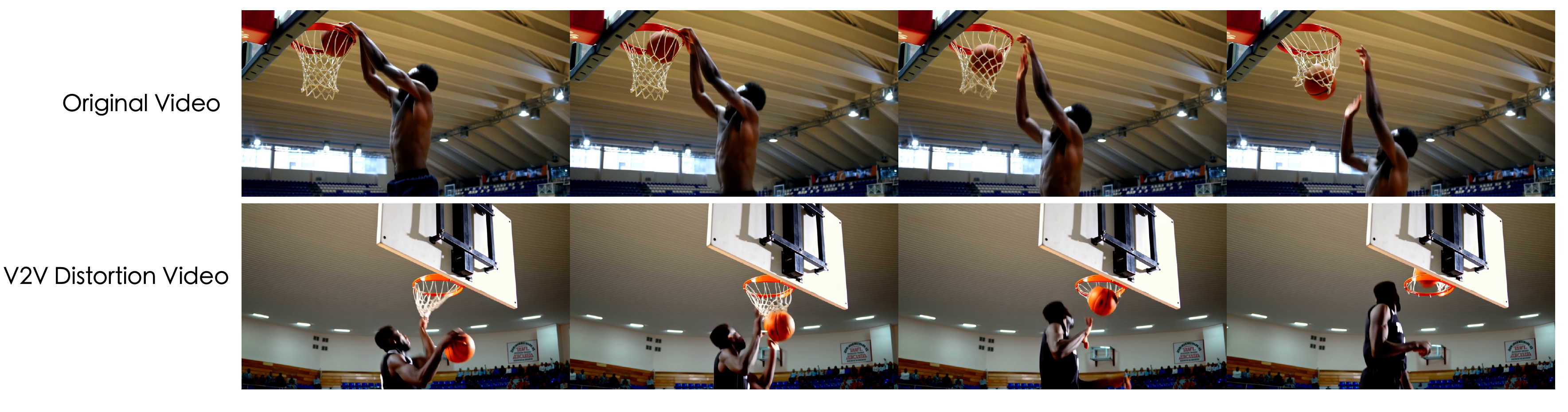}
    \captionsetup{justification=raggedright, singlelinecheck=false}
    \caption{An example of T2V distortion: The generated video shows the basketball rising upward instead of falling downward (violating gravity).}
    \label{fig:vert2a}
  \end{subfigure}
  
  \captionsetup{justification=raggedright, singlelinecheck=false}
  \caption{Examples of various distortion types created by our progressive distortion creation process.}
  \label{fig:vertical-compare}
\end{figure}

\paragraph{Preference Data Annotated by Human} 

Through rigorous analysis of motion artifacts in generated videos, we establish a systematic taxonomy of prevalent failure modes: excessive/insufficient motion amplitude, subject distortion, local detail corruption, physical law violations, and unnatural motion. In addition, we record the prompts corresponding to these failure modes and generate the same type of prompts by LLMs. Those generated prompts are diverse, ranging from human and animal interaction to object movement, including all the above types of motion failures. Then, each prompt is used to generate four samples with a history checkpoint pool of our pre-trained models.

After sample collection phase, samples for the same prompt were systematically paired into sample pairs. Professional human annotators are invited to rate the preference of these sample pairs. Our annotation pipeline follows two main steps: 1) \emph{Data filtering}: Samples will be excluded under two conditions: First, content/quality mismatch - if the two samples describe different textual contents or exhibit significant visual quality discrepancies, to ensure focus on motion quality analysis; Second, Annotation criteria failure - if either sample in the pair fails to meet the three criteria -- clarity of the main subject, sufficient subject size within the image frame, or simplicity of the background composition. From our experience, this process will drop almost 80\% of data pair before proceeding to further annotation. 2) \emph{Preference selection}: Human Annotators assign  one of three label (Better/Worse/Tie) to each sample pair according to the motion quality criteria. The details of motion criteria for human annotation are listed in Tab~\ref{tab:motion_quality_assessment}, which provides descriptions of all failure types in motion quality. Each failure type is assigned a weighted score, and the overall scores of both videos are calculated to enable comparison between them.
\paragraph{Preference Data Automatic Generation}
The resource-intensive nature of human annotation under our stringent quality criteria significantly constrained dataset scale. To augment the preference datasets, we design a automatic preference data generation pipeline, comprising two core steps: 
\paragraph{1) Ground Truth data collection}
We use the generated prompts to query our existing dataset to obtain the similar prompts by calculating cosine similarities between their CLIP features\cite{radford2021learning}. The curated ground-truth reference videos associated with semantically matched prompts serve as chosen samples. The rejected samples are generated through the following step to form preference pairs.
\paragraph{2) Progressively Distortion Creation}
The basic observation is that state-of-the-art video generation models still fall short in motion quality compared to real videos. We address this by deliberately adding controllable distortions to real videos, creating systematic simulations of motion flaws. Each real video comes with a text caption (describing content) and its first frame (static reference), enabling dynamic defect analysis while preserving visual structure. We create three variants of corrupted samples: V2V: Direct inversion of noisy latent (lowest distortion); I2V: Reconstruction using first frame guidance (medium distortion); T2V: Regeneration from text description (highest distortion). Also, we use different generative models (\cite{wan2025,kong2024hunyuanvideo,yang2024cogvideox}) and model parameters (e.g., timestep) to construct different levels of motion quality, while maintaining sample diversity. Fig.~\ref{fig:vertical-compare} shows three cases constructed using our process automation.

Beyond standard procedures, our research has explored innovative techniques to induce specific video quality issues. We can manipulate frame sampling rates in the temporal domain, increasing or decreasing them to create excessive or insufficient motion amplitude effects, or alternating rates for erratic motion. Using the Tea-Cache~\cite{liu2024timestep} method, we can tune parameters and inject noise to corrupt local details in video frames. For scenarios like cars driving or birds flying, we create pairs by playing videos backwards, challenging models to distinguish correct from incorrect physical motions. These methods are highly effective in simulating various bad cases in video generation. They can accurately replicate scenarios such as abnormal motion, local detail loss, and physics, which defying actions that might occur during the video generation process.

\paragraph{Reward model training}
Following VideoAlign~\cite{liu2025improving}, we implement our motion quality reward with Qwen2.5-VL-7B-Instruct~\cite{Qwen2.5-VL}. The training data comes from the above data collecting process, forming a total of $30k$ sample pairs. Since the motion quality is context-agnostic, sample pairs do not include the prompts. The model is trained with BradleyTerry model with ties (BTT)~\cite{rao1967ties}, an extension of BT that accounts for tied preferences: $\mathcal{L} = -\sum_{(i,j)} \left[ y_{i>j} \ln P(i>j) + y_{i<j} \ln P(i<j) + y_{i=j} \ln P(i=j\right)]$. Where $i>j$, $i<j$, $i=j$ mean that sample $i$ is better than/worse than/equal to sample $j$ in the sample pairs.

\paragraph{DPO training}
We apply the direct preference optimization for flow (Flow-DPO) from~\cite{liu2025improving} to improve the motion quality of our generative model. The loss function can be defined as:
\begin{equation}
\mathcal{L}_{\text{DPO}} = -\frac{1}{N} \sum_{i=1}^N \cdot \log\sigma\left( -\frac{\beta}{2} \Big[ \underbrace{(L_{\text{model}}^w - L_{\text{model}}^l)}_{\Delta_{\text{model}}} - \underbrace{(L_{\text{ref}}^w - L_{\text{ref}}^l)}_{\Delta_{\text{ref}}} \Big] \right)
\end{equation}
\begin{align*}
\text{wherein:} & \\
& L_{\text{model}}^w = \frac{1}{2}\|\mathbf{\hat{y}}_{\text{model}}^w - \mathbf{y}\|_2^2, \quad 
  L_{\text{model}}^l = \frac{1}{2}\|\mathbf{\hat{y}}_{\text{model}}^l - \mathbf{y}\|_2^2 \\
& L_{\text{ref}}^w = \frac{1}{2}\|\mathbf{\hat{y}}_{\text{ref}}^w - \mathbf{y}\|_2^2, \quad 
  L_{\text{ref}}^l = \frac{1}{2}\|\mathbf{\hat{y}}_{\text{ref}}^l - \mathbf{y}\|_2^2 \\
& \Delta_{\text{model}} = L_{\text{model}}^w - L_{\text{model}}^l, \quad 
  \Delta_{\text{ref}} = L_{\text{ref}}^w - L_{\text{ref}}^l
\end{align*}
Where the $\beta$ is a temperature coefficient. $\mathbf{\hat{y}}_{\text{model}}^{w/l}$ is the current model predictions for chosen/rejected samples. And $\mathbf{\hat{y}}_{\text{ref}}^{w/l}$ are the reference model predictions for chosen/rejected samples. 

To collect these training samples, we construct two types of prompt sets: concept-balanced prompts (for diversity) and motion-specific prompts (for motion quality). Each prompt is used to generate 8 videos using our generative model. Then we use the motion quality reward model to rank the videos and choose the best video and worst video, forming a sample triplet (chosen video, rejected video, prompt). Note that our DPO training is conducted in stages. When the model begins to easily distinguish between chosen and rejected samples (indicating performance plateaus), we refresh the reference model with the latest iteration. The updated reference model then generates new data, which is ranked by the reward model to form training data for the next stage. Each stage needs $20k$ training data, And we conduct a total of 3 stages of DPO training.

\subsubsection{Diffusion Forcing}
\label{subsec:diffusion_forcing}
In this section, we introduce the Diffusion Forcing Transformer, which unlocks our model’s ability to generate long videos. Diffusion Forcing~\cite{chen2024diffusionforcingnexttokenprediction} is a training and sampling strategy where each token is assigned an independent noise level. This allows tokens to be denoised according to arbitrary, per-token schedules using the trained model. Conceptually, this approach functions as a form of partial masking: a token with zero noise is fully unmasked, while complete noise fully masks it. Diffusion Forcing trains the model to “unmask” any combination of variably noised tokens, using the cleaner tokens as conditional information to guide the recovery of noisy ones. Building on this, our Diffusion Forcing Transformer can extend video generation indefinitely based on the last frames of the previous segment. 
Note that the synchronous full sequence diffusion is a special case of Diffusion Forcing, where all tokens share the same noise level. This relationship allows us to fine-tune the Diffusion Forcing Transformer from a full-sequence diffusion model.

Inspired by AR-Diffusion~\cite{sun2025ar}, we utilize the Frame-oriented Probability Propagation (FoPP) timestep scheduler for Diffusion Forcing Training. The process involves the following steps:

\begin{enumerate}
    \item \textbf{Uniform Sampling}: First, we uniformly sample a frame index \( f \sim U(1, F) \) and a corresponding timestep \( t \sim U(1, T) \). This ensures that timesteps are evenly distributed across all video frames.
    
    \item \textbf{Dynamic Programming for Probability Propagation}: Using dynamic programming, we calculate the probabilities for timesteps of frames before and after the sampled frame \( f \), conditioned on \( t_f = t \).
    
    \item \textbf{Definition of Transition Equation}: We define \( d^s_{i,j} \) as the count of valid timestep sequences starting at frame \( i \) with timestep \( j \), under the non-decreasing constraint. We compute \( d^s_{i,j} \) using the transition equation:
    \[
    d_{i,j} = d_{i,j-1} + d_{i-1,j}
    \]
    with boundary conditions \( d_{*, T} = 1 \) and \( d_{F, *} = 1 \).
    
    \item \textbf{Visit Probability Calculation}: For frames after \( f \), the probability of visiting timestep \( k \) is:
    \[
    \frac{d^s_{i,k}}{\sum_{j=K}^T d^s_{i,j}}
    \]
    Similarly, for frames before \( f \), we define \( d^e_{i,j} \) and calculate the probability as:
    \[
    \frac{d^e_{i,k}}{\sum_{j=1}^K d^e_{i,j}}
    \]
    
    \item \textbf{Timestep Sampling}: Finally, timesteps for previous or subsequent frames are sampled one by one based on the calculated probabilities.
\end{enumerate}

During inference, we adapt an Adaptive Difference (AD) timestep scheduler~\cite{sun2025ar} that supports adaptive video generation, accommodating both asynchronous auto-regressive and synchronous generation.

The AD scheduler treats the timestep difference between neighboring frames as an adaptive variable \(s\). For consecutive frames with timesteps \(t_i\) and \(t_{i-1}\), the condition is:
\[
t_i = 
\begin{cases} 
t_i + 1, & \text{if } i = 1 \text{ or } t_{i-1} = 0, \\
\min(t_{i-1} + s, T), & \text{if } t_{i-1} > 0
\end{cases}
\]

When the previous frame is none or clean, the current frame focuses on self-denoising. Otherwise, it denoises with a timestep difference of \(s\) from the previous frame. Notably, synchronous diffusion (\(s = 0\)) and auto-regressive generation (\(s = T\)) are special cases. A smaller \(s\) yields more similar neighboring frames, while a larger \(s\) increases content variability. 

Our conditioning mechanism enables auto-regressive frames generation by leveraging cleaner historical samples as conditions. In this framework, the information flow is inherently directional: noisy samples rely on preceding history to ensure consistency. This directional nature implies that bidirectional attention is unnecessary and can be replaced with more efficient causal attention. 
After training the Diffusion Forcing Transformer with bidirectional attention, one can fine-tune the model with context-causal attention for enhanced efficiency. During inference, this architecture enables caching of K, V features from historical samples, eliminating redundant computations and significantly reducing computational overhead.

\subsubsection{High-quality Supervised Fine-Tuning (SFT)}
\label{subsec:SFT}
We implement two sequential high-quality supervised fine-tuning (SFT) stages at 540p and 720p resolutions respectively, with the initial SFT phase conducted immediately after pretraining but prior to reinforcement learning (RL) stage. This first-stage SFT serves as a conceptual equilibrium trainer, building upon the foundation model's pretraining outcomes that utilized only fps24 video data, while strategically removing FPS embedding components to streamline the architecture. Trained with the high-quality concept-balanced samples as detailed in Section~\ref{para:databalnce}, this phase establishes optimized initialization parameters for subsequent training processes. Following this, we execute a secondary high-resolution SFT at 720p after completing the diffusion forcing stage, incorporating identical loss formulations and the higher-quality concept-balanced datasets by the manually filter. This final refinement phase focuses on resolution increase such that the overall video quality will be further enhanced.
\section{Infrastructure}
\label{subsec:infrastructure}
In this section, we introduce the infrastructure optimizations during the training and inference stages.
\subsection{Training optimization}
The training optimization focuses on ensuring efficient and robust training, including \emph{memory optimization}, \emph{training stability}, and \emph{parallel strategy}, which are elaborated in the following paragraphs.
\paragraph{Memory Optimization}
The attention block's fp32 memory-bound operations dominate GPU memory usage. We address this through efficient operator fusion, reducing kernel launch overhead while optimizing memory access and utilization for improved performance. Gradient checkpointing (GC) minimizes memory by storing only transformer block inputs in fp32; converting these to bf16 cuts memory by 50\% with negligible accuracy impact. Activation offloading further saves GPU memory by asynchronously moving temporary tensors to CPU, preserving throughput. However, due to shared CPU memory across 8 GPUs and limited computation overlap with excessive offloading, we strategically combine GC with selective activation offloading for optimal efficiency.
\paragraph{Training Stability}
We propose an intelligent self-healing framework that implements autonomic fault recovery through three-phase remediation: real-time detection and isolation of compromised nodes, dynamic resource reallocation using standby computing units, and task migration with checkpoint restoration to ensure uninterrupted model training.

\paragraph{Parallel Strategy}
We pre-compute the results of VAE and text encoder. Using FSDP to distributively store DiT's weights and optimizer states across all nodes to address the GPU memory pressure caused by the large model size.
When training at 720p resolution, due to large temporary tensors, we encounter severe GPU memory fragmentation issues, triggering \verb|torch.empty_cache()| even when memory is still sufficient. Therefore, we use Sequence Parallel \cite{jacobs2023deepspeedulyssesoptimizationsenabling} to address the memory pressure caused by activations.

\subsection{Inference optimization}

The key goal of inference optimization is to reduce video generation latency without compromising quality. While diffusion-based model succeed in producing high-fidelity video, its requires multi-step samplings
during inference, which is typically 30 to 50 steps that can take more than 5 minutes for 5 seconds video. In our actual deployment, we achieved optimization through \emph{VRAM optimization}, \emph{quantization}, \emph{multi-GPU parallel}, and \emph{Distillation}.
\paragraph{VRAM optimization} Our deployment leverages RTX 4090 GPUs (24GB VRAM) to serve the 14B-parameter model. By combining FP8 quantization with parameter-level offloading techniques, we successfully enable 720p video generation while maintaining full model capabilities on a single GPU instance.
\paragraph{Quantization} Our analysis identifies the attention  and linear layers as the primary computational bottlenecks in DiTs. To optimize performance, we implement FP8 quantization across the architecture. Specifically, we apply FP8 dynamic quantization combined with FP8 GEMM acceleration to linear layers, achieving \emph{1.10$\times$} speedup compared to the bf16 baseline on RTX 4090 hardware. For attention operations, we deploy sageAttn2-8bit\cite{zhang2025sageattention2efficientattentionthorough}, which delivers \emph{1.30$\times$} faster inference than the bf16 implementation on the same RTX 4090 platform.
\paragraph{Parallel strategy} To accelerate single-video generation, we utilize three key parallelization strategies: Content Parallel, CFG Parallel, and VAE Parallel. In real-world deployment, this approach reduces overall latency by \emph{1.8$\times$} when scaling from 4 to 8 RTX 4090 GPUs. 
\paragraph{Distillation} To accelerate the video generation, we employ DMD distillation technique~\cite{yin2024onestepdiffusiondistributionmatching,yin2024improveddistributionmatchingdistillation}. We remove the regression loss and use high-quality video data instead of pure noise as the input to the student generator to accelerate model convergence. Besides, we also adopt the two time-scale update rule to ensure the fake score generator tracks the student generator's output distribution and the multi-step schedule from DMD. Similarly, as demonstrated in the formulation, the gradients are applied to update the student generator $G$. 
\begin{equation*}
\nabla_{\theta} D_{KL} \simeq \mathbb{E}_{t,x} \left[\left( s_{\text{fake}}(x,t) - s_{\text{real}}(x,t) \right) \frac{dG}{d\theta} \right]
\end{equation*}
where $x$ represents the video generated by the student generator, $s_{fake}$ and $s_{real}$ denote the evaluation scores produced by the fake score generator and real score generator, respectively. We set the update ratio of fake score generator and student generator as 5 and use 4-step generator with a specified schedule tuned for flow matching framework. During the distillation stage, we find a small learning rate combined with a larger batch size is very important to stablize the trainining. Through the aforementioned distillation process, we can significantly reduce the time required for video generation.

\section{Performance}
\label{subsec:performance}
To comprehensively evaluate our proposed method, we construct the SkyReels-Bench for human assessment and leverage the open-source V-Bench for automated evaluation. This allows us to compare our model with the state-of-the-art (SOTA) baselines, including both open-source and proprietary models.

\subsection{SkyReels-Bench}
For human evaluation, we design SkyReels-Bench with 1,020 text prompts, systematically assessing three dimensions: Instruction Adherence, Motion Quality, Consistency and Visual Quality. This benchmark is designed to evaluate both text-to-video (T2V) and image-to-video (I2V) generation models, providing comprehensive assessment across different generation paradigms.

\paragraph{Instruction Adherence}
 Evaluating how well the generated video follows the provided text prompt.
1) \emph{Motion instruction adherence}: Accuracy in executing specified actions or movements. 
2) \emph{Subject instruction adherence}: Correct representation of described subjects and attributes.
3) \emph{Spatial relationships}: Proper positioning and interaction between subjects.
4) \emph{Shot adherence}: Correct implementation of specified shot types (close-up, wide, etc.). 
5) \emph{Expression adherence}: Accurate portrayal of emotional states and facial expressions.
6) \emph{Camera motion adherence:}: Proper execution of camera movements (pan, tilt, zoom, etc.).
7) \emph{Hallucination}: Absence of content not specified in the prompt

\paragraph{Motion Quality} Assessing the temporal dynamics of subjects in the video.
1) \emph{Motion dynamism}: Diversity and expressiveness of movements.
2) \emph{Fluidity and stability}: Smoothness of motion without jitter or discontinuities.
3) \emph{Physical plausibility}: Adherence to natural physics and realistic movement patterns.

\paragraph{Consistency} Measuring coherence across video frames.
1) \emph{Subject consistency}: Stable appearance of main subjects throughout the video.
2) \emph{Scene consistency}: Coherent background, location, and environmental elements.
For image-to-video (I2V) models, we additionally evaluate:
3) \emph{First-frame fidelity}: Consistency between the generated video and the provided input image, including preservation of color palette, maintenance of subject identity and continuity of scene elements established in the first frame.

\paragraph{Visual Quality} Evaluating the spatial fidelity of the generated content.
1) \emph{Visual clarity}: Sharpness and definition of visual elements.
2) \emph{Color accuracy}: Appropriate color balance without oversaturation.
3) \emph{Structural integrity}: Absence of distortions or corruptions in subjects and backgrounds.

This comprehensive evaluation framework allows us to systematically compare video generation capabilities across different models and identify specific strengths and weaknesses in various aspects of video quality.

For evaluation, a panel of 20 professional evaluators assess each dimension using a 1-5 scale, with the following criteria as shown in Table~\ref{tab:rating_scale}:
\begin{table}[htbp]
\centering
\caption{Video Quality Assessment Rating Scale}
\label{tab:rating_scale}
\begin{tabular}{clp{10cm}}
\toprule
\textbf{Score} & \textbf{Label} & \textbf{Assessment Criteria} \\
\midrule
1 & Fail & Complete failure to meet evaluation criteria \\
2 & Marginal & Partial compliance with significant deficiencies \\
3 & Adequate & Basic compliance with non-critical flaws \\
4 & Proficient & Full satisfaction of all requirements \\
5 & Excellent & Exceptional quality exceeding baseline requirements \\
\bottomrule
\end{tabular}
\end{table}

The final results are summarized in Table~\ref{tab:skyreels-bench-t2v}. The evaluation demonstrates that our model achieves significant advancements in instruction adherence compared to baseline methods, while maintaining competitive performance in motion without sacrificing the consistency. To ensure fairness, all models are evaluated under default settings with consistent resolutions, and no post-generation filtering is applied. Detailed scoring guidelines for each criterion can be found in the Appendix~\ref{app:skyreels-bench-detailed}.

\begin{table}[htbp]
\centering
\setlength{\tabcolsep}{0pt}
\begin{tabular*}{\textwidth}{@{\extracolsep{\fill}}lccccc@{}}
\toprule
\textbf{Model Name} & \textbf{Average} & \textbf{Instruction Adherence} & \textbf{Consistency} & \textbf{Visual Quality} & \textbf{Motion Quality} \\
\midrule
Runway-Gen3 Alpha \cite{runwayml2025gen3alpha} & 2.53 & 2.19 & 2.57 & 3.23 & 2.11 \\
HunyuanVideo-13B \cite{kong2024hunyuanvideo} & 2.82 & 2.64 & 2.81 & 3.20 & 2.61 \\
Kling-1.6 STD Mode \cite{kuaishou2024kling} & 2.99 & 2.77 & 3.05 & 3.39 & \textbf{2.76} \\
Hailuo-01 \cite{minimax2024hailuo} & 3.0 & 2.8 & 3.08 & 3.29 & 2.74 \\
Wan2.1-14B \cite{wan2025} & 3.12 & 2.91 & 3.31 & \textbf{3.54} & 2.71 \\
\textbf{SkyReels-V2} & \textbf{3.14} & \textbf{3.15} & \textbf{3.35} & 3.34 & 2.74 \\
\bottomrule
\end{tabular*}
\vspace{4pt}
\caption{\textbf{Text-to-Video (T2V) Model Performance on SkyReels-Bench.} Evaluation conducted on a 1-5 scale across multiple dimensions, with higher scores indicating better performance.}
\label{tab:skyreels-bench-t2v}
\end{table}

\subsection{Model Benchmarking and Leaderboard}
To objectively compare \textbf{SkyReels-V2} against other leading open-source video generation models, we conduct comprehensive evaluations using the public benchmark VBench1.0~\cite{huang2023vbench}. 

Our evaluation specifically leverages the benchmark's longer version prompt. For fair comparison with baseline models, we strictly follow their recommended setting for inference. Meanwhile, our model employs 50 inference steps and guidance scale of 6 during generation, aligning with the common practice.

\begin{table}[htbp]
\centering
\begin{tabular}{l@{\hspace{2em}}|@{\hspace{2em}}c@{\hspace{2em}}c@{\hspace{2em}}c}
\toprule
\textbf{Model} & \textbf{Total Score} & \textbf{Quality Score} & \textbf{Semantic Score} \\
\midrule
CogVideoX1.5-5B \cite{yang2024cogvideox} & 80.3 \% & 80.9 \% & 77.9 \% \\
OpenSora-2.0 \cite{opensora2} & 81.5 \% & 82.1 \% & 78.2 \% \\
HunyuanVideo-13B \cite{kong2024hunyuanvideo} & 82.7 \% & 84.4 \% & 76.2 \% \\
Wan2.1-14B \cite{wan2025} & 83.7 \% & 84.2 \% & \textbf{81.4 \%}  \\
\textbf{SkyReels-V2} & \textbf{83.9 \%} & \textbf{84.7 \%} & 80.8 \% \\
\bottomrule
\end{tabular}
\vspace{4pt}
\caption{\textbf{Text-to-Video (T2V) Model Performance on Vbench1.0's long prompt version}}
\label{tab:comparison-vbench}
\end{table}

The VBench results (Table~\ref{tab:comparison-vbench}) demonstrate that \textbf{SkyReels-V2} outperforms all baseline models including HunyuanVideo-13B and Wan2.1-14B, With the highest total score (83.9\%) and quality score (84.7\%). In this evaluation, the semantic score is slightly lower than Wan2.1-14B, while we outperform Wan2.1-14B in the previous human evaluation, with the primary gap attributed to V-Bench's insufficient evaluation of shot-scenario semantic adherence.
\begin{figure}[t]
\centering
\includegraphics[width=\linewidth]{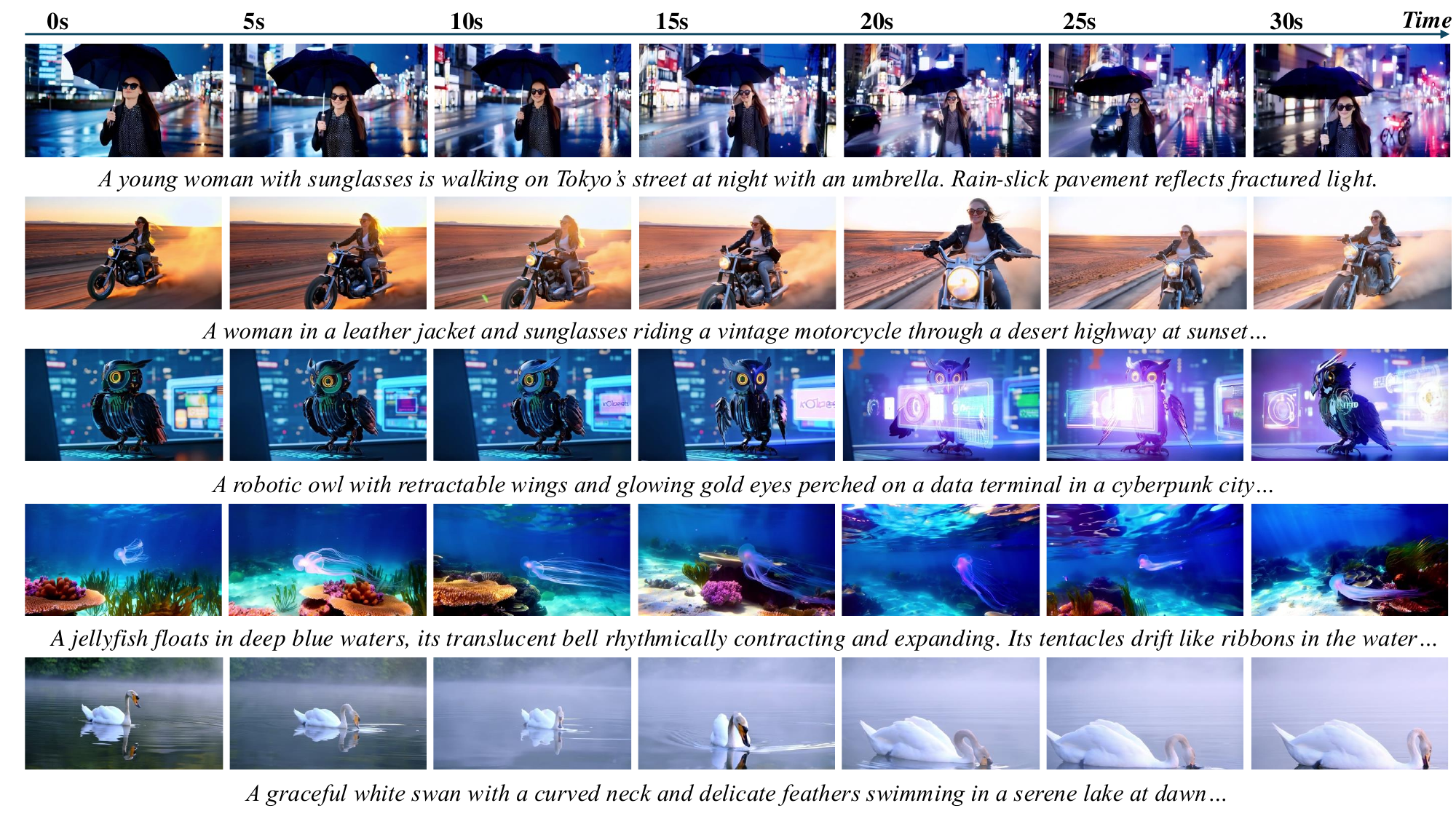}
\caption{Examples of ultra-long video generation using single prompts with our \textbf{SkyReels-V2} model.}
\label{fig:30s_single}
\end{figure}

\begin{figure}[t]
\centering
\includegraphics[width=\linewidth]{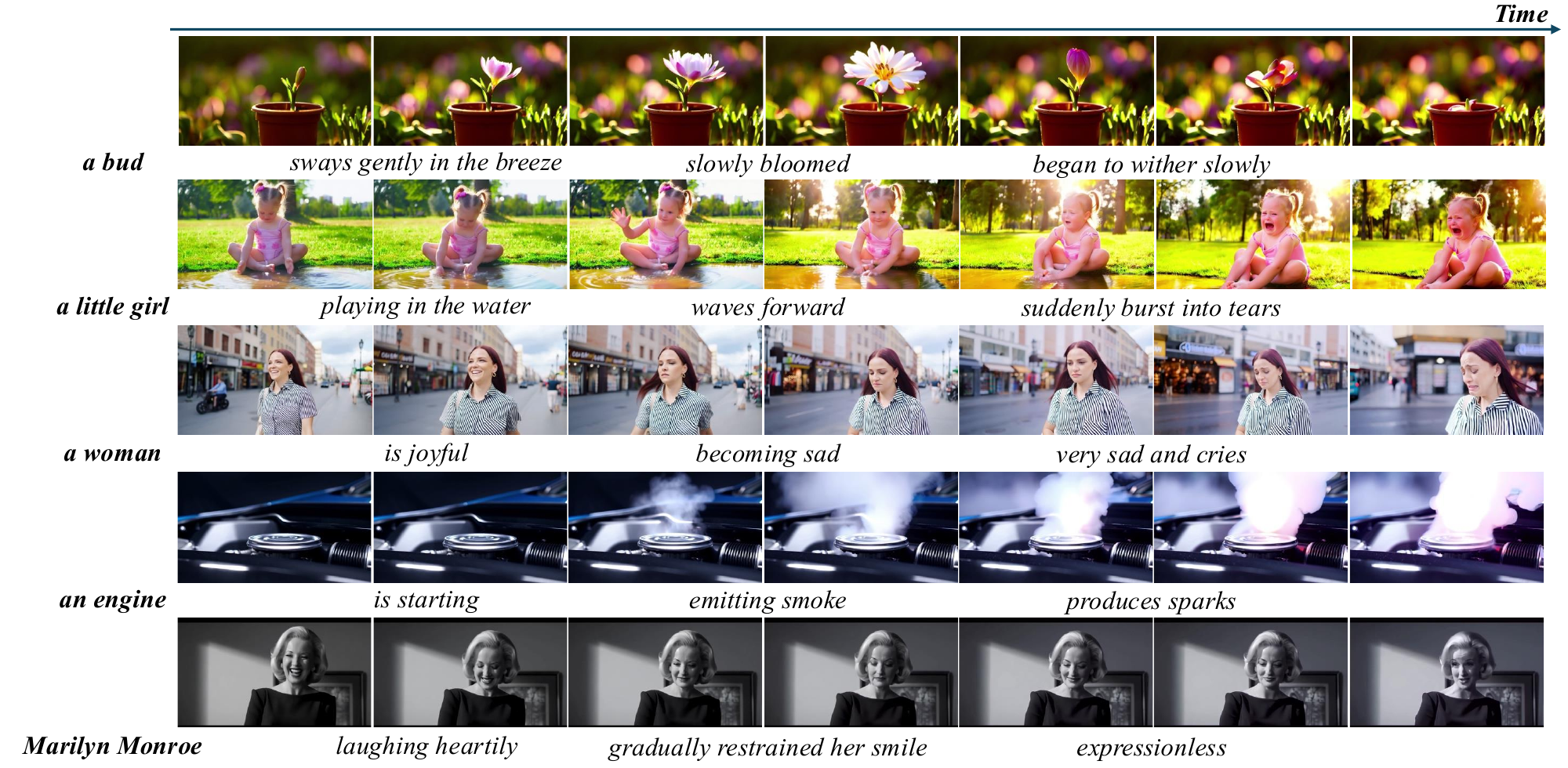}
\caption{Examples of ultra-long video generation using sequential prompts with our \textbf{SkyReels-V2} model.}
\label{fig:30s_seq}
\end{figure}

\section{Application}
\label{subsec:application}
\subsection{Story Generation}
Our trained diffusion forcing transformer enables the generation of ultra-long videos with theoretically unlimited extension. 
The model generates long videos using a sliding window approach. It conditions on the last $f_{prev}$ frames and the text prompt to generate the next $f_{new}$ frames, except for the first iteration, where it relies solely on the text prompt.
However, extending the video length can lead to error accumulation over time. To mitigate this issue, we employ a stabilization technique where previously generated frames are marked with slight noise level. This approach prevents error accumulation and further stabilizes the long rollout process. In Figure~\ref{fig:30s_single}, we illustrate examples of extending long shot videos to durations exceeding 30 seconds, demonstrating the capability to enhance temporal length while preserving visual coherence.

Our model not only supports mere temporal extension but is also able in generating long shots with compelling story narratives. By leveraging a sequence of narrative text prompts, we can orchestrate a cohesive visual narrative that spans multiple actions while maintaining visual consistency throughout the video. This capability ensures smooth transitions between scenes, allowing for dynamic storytelling without compromising the integrity of the visual elements. The model's ability to seamlessly integrate narrative text prompts enables the creation of extended video content that is both engaging and visually harmonious, making it ideal for applications requiring complex, multi-action sequences. In Figure~\ref{fig:30s_seq}, we illustrate examples where users can manipulate attributes such as the actions of the little girl, the expressions of the woman, and the status of the engine through sequential text prompts.




\subsection{Image-to-Video Synthesis}
There are two approaches to develop image-to-video (I2V) models under our frameworks: 1) \emph{Fine-Tuning full-sequence Text-to-Video (T2V) diffusion Models}~(\textbf{SkyReels-V2-I2V})}: Following Wan 2.1's I2V implementation, we extend T2V architectures by injecting the first reference frame as an image condition. The input image is padded to match the target video length, then processed through a VAE encoder to obtain image latents. These latents are concatenated with noise latents and 4 binary mask channels (1 for the reference frame, 0 for subsequent frames), enabling the model to leverage the reference frame for subsequent generation. To preserve original T2V capabilities during adaptation, we apply zero-initialization to newly added convolutional layers and specifically to the image context to value projections in cross attention, while other new components (such as image context to key projections) use random initialization, minimizing abrupt performance shifts during fine-tuning. Besides, the I2V training leverages I2V-specific prompts generated through the captioning framework described in Section \ref{sec:video_captioner}. Remarkably, this approach achieves competitive results with only 10,000 training iterations on 384 GPUs. 2) \emph{Text-to-Video (T2V) Diffusion Forcing model with first-frame Conditioning}~(\textbf{SkyReels-V2-DF}): Our alternative method directly utilizes the diffusion framework's conditioning mechanism by feeding the first frame as a clean reference condition. This bypasses explicit model retraining while maintaining temporal consistency through latent space constraints. We evaluate SkyReels-V2 against leading open-source and closed-source image-to-video models using the SkyReels-Bench evaluation suite (Table~\ref{tab:skyreels-bench-i2v}). Our results demonstrate that both \textbf{SkyReels-V2-I2V}~(3.29) and \textbf{SkyReels-V2-DF}~(3.24) achieve state-of-the-art performance among open-source models, significantly outperforming HunyuanVideo-13B~(2.84) \cite{kong2024hunyuanvideo} and Wan2.1-14B~(2.85) \cite{wan2025} across all quality dimensions. With an average score of 3.29, SkyReels-V2-I2V demonstrates comparable performance to proprietary models Kling-1.6 \cite{kuaishou2024kling} (3.4) and Runway-Gen4 \cite{runwayml2024gen4} (3.39). Based on these promising results, we publicly release our SkyReels-V2-I2V model to advance community research in Image-to-video synthesis.

\begin{table}[htbp]
\centering
\setlength{\tabcolsep}{0pt}
\begin{tabular*}{\textwidth}{@{\extracolsep{\fill}}lccccc@{}}
\toprule
\textbf{Model} & \textbf{Average} & \textbf{Instruction Adherence} & \textbf{Consistency} & \textbf{Visual Quality} & \textbf{Motion Quality} \\
\midrule
HunyuanVideo-13B \cite{kong2024hunyuanvideo} & 2.84 & 2.97 & 2.95 & 2.87 & 2.56 \\
Wan2.1-14B \cite{wan2025} & 2.85 & 3.10 & 2.81 & 3.00 & 2.48 \\
Hailuo-01 \cite{minimax2024hailuo} & 3.05 & 3.31 & 2.58 & 3.55 & 2.74 \\
Kling-1.6 Pro Mode \cite{kuaishou2024kling}& 3.4 & 3.56 & 3.03 & 3.58 & 3.41 \\
Runway-Gen4 \cite{runwayml2024gen4} & 3.39 & 3.75 & 3.2 & 3.4 & 3.37 \\
\textbf{SkyReels-V2-DF} & 3.24 & 3.64 & 3.21 & 3.18 & 2.93 \\
\textbf{SkyReels-V2-I2V} & 3.29 & 3.42 & 3.18 & 3.56 & 3.01 \\
\bottomrule
\end{tabular*}
\vspace{4pt}
\caption{\textbf{Image-to-Video (I2V) Model Performance on SkyReels-Bench.} Evaluation conducted on a 1-5 scale across multiple dimensions, with higher scores indicating better performance.}
\label{tab:skyreels-bench-i2v}
\end{table}

\subsection{Camera Director}
Although SkyCaptioner-V1 demonstrates robust performance in annotating camera motions, we observe that while it achieves balanced subject distribution, the inherent imbalance in camera-motion data poses challenges for further optimization of cinematography parameters. To address this limitation, we specifically curate approximately 1 million samples from the supervised fine-tuning (SFT) dataset, ensuring a balanced representation of both basic camera motions and their common combinations. Building upon this enhanced dataset, we conduct fine-tuning experiments on our image-to-video generation model using 384 GPUs over 3,000 iterations. This dedicates training regimen resulted in marked enhancement of cinematographic effects, particularly in the fluidity and diversity of camera motions.
\subsection{Elements-to-Video Generation}
\begin{figure}[H]
  \includegraphics[width=\textwidth]{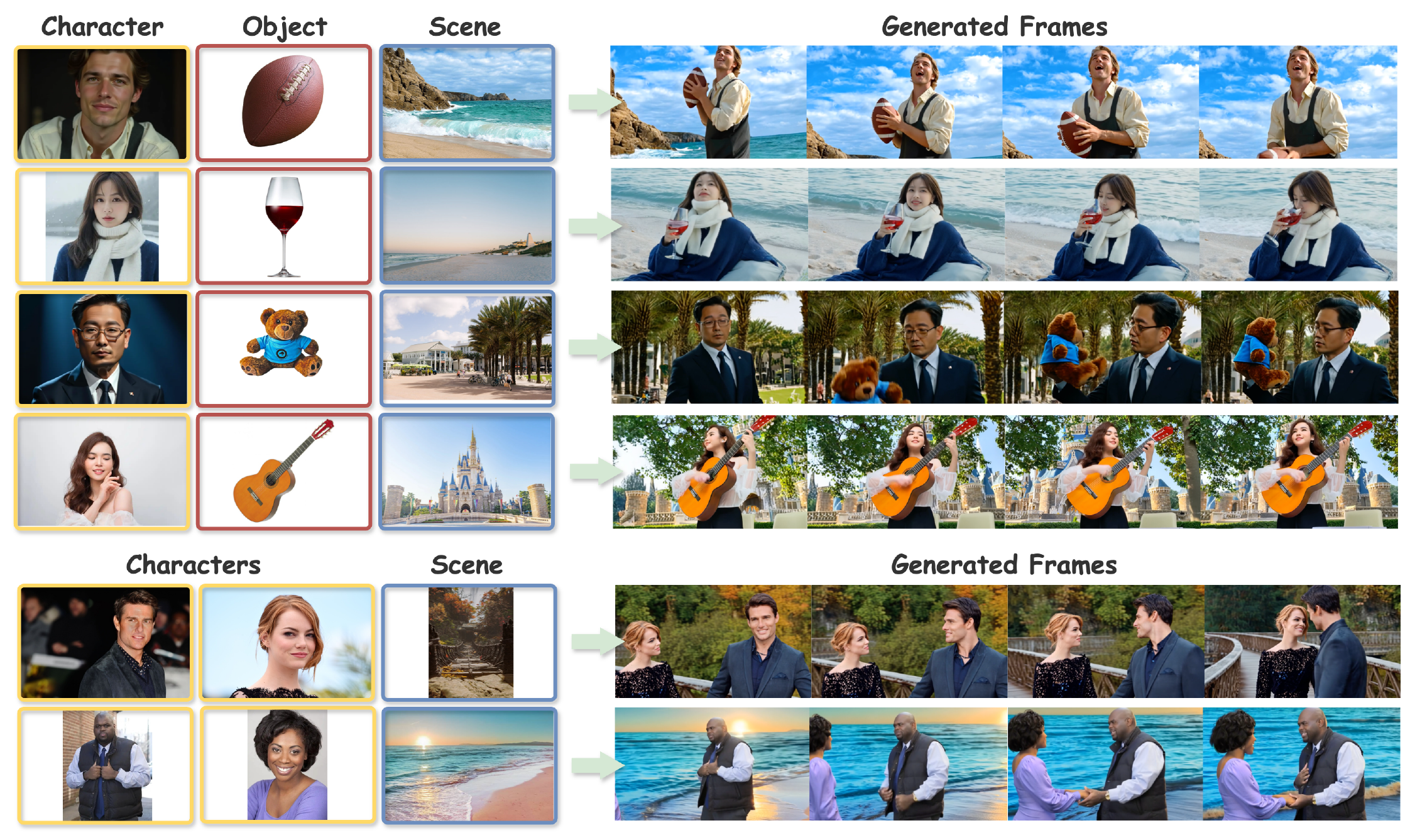}
  \caption{\textbf{Examples of \emph{elements-to-video} results from our proposed \textbf{SkyReels-A2} model.} Given reference with multiple images and textual prompt, our method can generate realistic and naturally composed videos while preserving specific identity consistent. 
  }
  \label{fig:a2case1} 
\end{figure}

Current video generation models mainly tackle two tasks: text-to-video (T2V) and image-to-video (I2V). T2V leverages text encoders like T5~\cite{raffel2019exploring} or CLIP~\cite{radford2021learning} to generate videos from textual prompts, but often suffers from inconsistency due to the randomness of the diffusion process. I2V, on the other hand, generates motion from a static image and optional text, yet is typically limited by over-dependence on the initial frame. In our previous work, we introduce an \emph{elements-to-video} (E2V) task and proposed \textbf{SkyReels-A2}\cite{fei2025skyreels}, a controllable video generation framework that composes arbitrary visual elements, such as characters, objects, and backgrounds, into coherent videos guided by textual prompts, while ensuring high fidelity to the reference images for each element. As shown in the Figure~\ref{fig:a2case1}, \textbf{SkyReels-A2} generates high-quality, temporally consistent videos with editable compositions of multiple visual elements.  Furthermore, we propose \textbf{A2-Bench}, a novel benchmark for comprehensively evaluating the E2V task, which exhibits statistically significant correlation with human subjective judgments.



In the future, we're planning to release a unified video generation framework that supports additional input modalities such as audio and pose. Building upon our previous work \textbf{SkyReels-A1}\cite{qiu2025skyreels} on audio-driven and pose-driven portrait animation, this enhanced framework will support richer and more diverse forms of input. By doing so, it aims to significantly broaden the scope of applications, including but not limited to short dramas, music videos, and virtual e-commerce content creation.

\section{Conclusion}
We propose the \textbf{SkyReels-V2} model, a novel framework capable of generating infinite-length videos while maintaining adherence to shot-scenario prompts, high video quality, and robust motion quality. Key improvements are achieved through the following advancements:
1) \emph{Prompt Adherence}: Enhanced through the \textbf{SkyCaptioner-V1} module, which leverages knowledge distillation from both general-purpose multi-modal large language models (MLLMs) and specialized shot-expert models to ensure precise alignment with input prompts. 2) \emph{Video Quality}: Significantly improved by leveraging diverse data sources and a multi-stage training pipeline, ensuring visually coherent and high-fidelity outputs. 3) \emph{Motion Quality}: Optimized via reinforcement learning post-training, supported by a semi-automated data production pipeline that refines dynamic motion consistency and fluidity. 4) \emph{Infinite-Length Generation}: Enabled by the diffusion-forcing framework, which allows seamless extension of video content without explicit length constraints.
Despite these advancements, the diffusion-forcing framework remains subject to error accumulation over extended generations, which currently limits the practical length of high-quality video outputs. Future work will focus on addressing this challenge to further enhance the model's scalability and reliability.
\section{Contributors}
We gratefully acknowledge all contributors for their dedicated efforts. The following lists recognize participants by their primary contribution roles:
\begin{itemize}
    \item \textbf{Project Sponsor:} Yahui Zhou

    \item \textbf{Project Leader:} Guibin Chen$^\dagger$ (guibin.chen@kunlun-inc.com)

    \item \textbf{Contributors:}
    \begin{itemize}
        \item \textit{Infrastructure:} Hao Zhang$^\dagger$, Weiming Xiong, Zhiheng Xu, Yuzhe Jin

        \item \textit{Data \& Captioning:} Mingyuan Fan$^\dagger$, Zheng Chen, Chengcheng Ma, Peng Zhao, Boyuan Xu

        \item \textit{Model Training:} Dixuan Lin$^\dagger$, Jiangping Yang$^\dagger$, Chunze Lin$^\dagger$, Junchen Zhu$^\dagger$, Sheng Chen, Wei Wang, Nuo Pang, Kang Kang, Yupeng Liang, Yubing Song, Di Qiu, Debang Li, Zhengcong Fei
    \end{itemize}

\end{itemize}
$^\dagger$ Indicates equally contributing authors.

\bibliographystyle{unsrt}  
\bibliography{template}  


\newpage
\appendix
\setcounter{table}{0} 
\renewcommand{\thetable}{A\arabic{table}} 
\setcounter{algorithm}{0}
\renewcommand{\thealgorithm}{A\arabic{algorithm}}

\section{SkyReels-Bench scoring guidelines}
\label{app:skyreels-bench-detailed}

Detailed scoring guidelines for SkyReels-Bench
\begin{table}[h]
\centering
\small
\renewcommand{\arraystretch}{1.2}
\setlength{\tabcolsep}{12pt}
\begin{tabular}{>{\raggedright\arraybackslash\bfseries}p{3cm}>{\raggedright\arraybackslash}p{12cm}}
\toprule
\multicolumn{2}{l}{\large\bfseries Evaluation Scoring Guidelines (1-5 scale)} \\
\midrule

 \textbf{Instruction Adherence} & 
\textbf{\underline{Score 1:}} Complete failure to follow instructions, with severe deviations in video theme and key elements from the prompt\\[0.5ex]
& \textbf{\underline{Score 2:}} Partial adherence with significant deviations; key elements or themes are somewhat present but incomplete and inaccurate\\[0.5ex]
& \textbf{\underline{Score 3:}} Basic adherence with complete representation of prompt content; main content and key elements align with instructions but may contain minor inaccuracies or omissions in details\\[0.5ex]
& \textbf{\underline{Score 4:}} High adherence with accurate representation of all key content and elements from the instructions without any deviations\\[0.5ex]
& \textbf{\underline{Score 5:}} Perfect adherence with extensions; video completely follows all aspects of instructions (theme, elements, style) and enhances them with appropriate extensions for superior results\\[1ex]
\midrule

\textbf{Consistency} & 
\textbf{\underline{Score 1:}} Complete inconsistency with severe discrepancies in subjects, style, and scenes; extreme frame-to-frame differences\\[0.5ex]
& \textbf{\underline{Score 2:}} Multiple significant inconsistencies that substantially impact overall video coherence\\[0.5ex]
& \textbf{\underline{Score 3:}} Partial inconsistencies limited to minor details (extremely localized areas); slight abnormalities in specific elements\\[0.5ex]
& \textbf{\underline{Score 4:}} Complete consistency with stable subjects, style, and scenes; minor imperfections only in non-primary elements that don't affect the overall viewing experience\\[0.5ex]
& \textbf{\underline{Score 5:}} Perfect consistency with extensions; all aspects of the video (subjects, style, scenes) are completely aligned with instructions, creating exceptional visual harmony\\[1ex]
\midrule

\textbf{Visual Quality} & 
\textbf{\underline{Score 1:}} Severe quality issues with significant blurriness, pixelation, or other visual artifacts making the video nearly unwatchable\\[0.5ex]
& \textbf{\underline{Score 2:}} Poor quality with noticeable blurriness and obvious issues; content is barely recognizable and significantly impacts viewing experience\\[0.5ex]
& \textbf{\underline{Score 3:}} Average quality with minor visual flaws such as slight blurriness or minimal noise; does not affect comprehension of main content\\[0.5ex]
& \textbf{\underline{Score 4:}} Good quality reaching normal viewing standards; clear imagery without notable flaws, presenting well across various devices\\[0.5ex]
& \textbf{\underline{Score 5:}} Perfect quality at professional standards; impeccable resolution, color, contrast, and detail presentation suitable for high-quality exhibitions and distribution\\
\midrule

\textbf{Motion Quality} & 
\textbf{\underline{Score 1:}} Severely flawed movements with extreme jerkiness and discontinuity, resulting in extremely poor viewing experience\\[0.5ex]
& \textbf{\underline{Score 2:}} Poor motion with obvious stuttering and disconnected transitions; movements appear unnatural with abrupt shifts between scenes\\[0.5ex]
& \textbf{\underline{Score 3:}} Adequate motion with occasional stuttering or discontinuities; generally follows movement rhythm without significantly affecting content comprehension\\[0.5ex]
& \textbf{\underline{Score 4:}} Good motion quality with smooth, natural movement throughout; no obvious stuttering and overall fluid viewing experience\\[0.5ex]
& \textbf{\underline{Score 5:}} Exceptional motion quality with perfect fluidity and naturalness; completely free of stuttering or discontinuities, achieving human-like movement without any AI artifacts\\[1ex]
\bottomrule

\end{tabular}
\vspace{4pt}
\caption{\textbf{Comprehensive Evaluation Rubric.} This scoring guideline was used by human evaluators to assess video generation models across four dimensions. Each dimension was scored on a 1-5 scale, where 1 indicates complete failure and 5 represents exceptional quality beyond base requirements.}
\label{tab:evaluation_guidelines}
\end{table}

\newpage
\section{Data Processing Pipeline}
\label{app:skyreels-data-processing}

The pseudo-code of finding largest interior rectangle in data processing is shown in following.

\begin{algorithm}
\caption{Pseudo-code of finding largest interior rectangle}

\begin{algorithmic}[1]
\Require A 0-1 $\text{matrix}$ $M\in\mathbb{R}^{m \times n}$
\Ensure The coordinate of largest interior rectangle $rect\_coords$

\State Initialize: $max\_area \gets 0$, $rect\_coords \gets (0, 0, 0, 0)$, $heights \gets [0] \times n$
\For{$i \gets 0$ \textbf{to} $m-1$}
    \For{$j \gets 0$ \textbf{to} $n-1$}
        \If{$\text{matrix}[i][j] = 1$}
            \State $heights[j] \gets heights[j] + 1$
        \Else
            \State $heights[j] \gets 0$
        \EndIf
    \EndFor
    
    \State $stack \gets \emptyset$, $left \gets [-1] \times n$
    \For{$j \gets 0$ \textbf{to} $n-1$}
        \While{$stack \neq \emptyset \text{ and } heights[j] \leq heights[\text{top}(stack)]$}
            \State $\text{pop}(stack)$
        \EndWhile
        \If{$stack \neq \emptyset$}
            \State $left[j] \gets \text{top}(stack)$
        \Else
            \State $left[j] \gets -1$
        \EndIf
        \State $\text{push}(stack, j)$
    \EndFor
    
    \State $stack \gets \emptyset$, $right \gets [n] \times n$
    \For{$j \gets n-1$ \textbf{downto} $0$}
        \While{$stack \neq \emptyset \text{ and } heights[j] \leq heights[\text{top}(stack)]$}
            \State $\text{pop}(stack)$
        \EndWhile
        \If{$stack \neq \emptyset$}
            \State $right[j] \gets \text{top}(stack)$
        \Else
            \State $right[j] \gets n$
        \EndIf
        \State $\text{push}(stack, j)$
    \EndFor
    
    \For{$j \gets 0$ \textbf{to} $n-1$}
        \State $height \gets heights[j]$
        \State $width \gets right[j] - left[j] - 1$
        \State $area \gets height \times width$
        \If{$area > max\_area$}
            \State $max\_area \gets area$
            \State $top \gets i - height + 1$
            \State $bottom \gets i$
            \State $left\_col \gets left[j] + 1$
            \State $right\_col \gets right[j] - 1$
            \State $rect\_coords \gets (top, left\_col, bottom, right\_col)$
        \EndIf
    \EndFor
\EndFor

\State \Return $rect\_coords$
\end{algorithmic}
\label{alg:max_inner_rectangle}
\end{algorithm}

\section{System Prompts of the SkyCaptioner-V1}
\label{app:skyreels-system-prompt}
The system prompt for the SkyCaptioner-V1 to generate structural caption is illustrated in Table~\ref{tab:captioner_comparison}. We also use the same system prompt for the baseline models during the evaluation.
\begin{tcolorbox}[
    breakable, 
    colback=white, 
    colframe=black!75, 
    title=System Prompt of Generating Structural Caption for Video, 
    fonttitle=\bfseries,
    width=\textwidth, 
  ]
I need you to generate a structured and detailed caption for the provided video. The structured output and the requirements for each field are as shown in the following JSON content: 

\{

\quad ``subjects": [
    
\qquad  \{

\qquad \quad ``TYPES": \{
            
\qquad \qquad     ``type": ``Main category (e.g., Human)",
                
\qquad \qquad     ``sub\_type": ``Sub-category (e.g., Man)"
                
\qquad \quad \}, 
        
\qquad \quad ``appearance": ``Main subject appearance description",
            
\qquad \quad ``action": ``Main subject action",
            
\qquad \quad ``expression": ``Main subject expression  (Only for human/animal categories, empty otherwise)",
            
\qquad \quad ``position": ``Subject position in the video (Can be relative position to other objects or spatial description)", 
            
\qquad \quad ``is\_main\_subject": true
            
\qquad  \}, 
        
\qquad  \{

\qquad \quad ``TYPES": \{
                
\qquad \qquad     ``type": ``Main category (e.g., Vehicles)",
                
\qquad \qquad     ``sub\_type": ``Sub-category (e.g., Ship)"

\qquad \quad \},
            
\qquad \quad ``appearance": ``Non\-main subject appearance description", 
            
\qquad \quad ``action": ``Non\-main subject action",
            
\qquad \quad ``expression": ``Non\-main subject expression (Only for human/animal categories, empty otherwise)",
            
\qquad \quad ``position": ``Position of non\-main subject 1",
            
\qquad \quad ``is\_main\_subject": false
            
\qquad  \}
        
\qquad  ],
    
\qquad  ``shot\_type": ``Shot type(Options: long\_shot/full\_shot/medium\_shot/close\_up/extreme\_close\_up/other)", 
    
\qquad  ``shot\_angle": ``Camera angle(Options: eye\_level/high\_angle/low\_angle/other)",
    
\qquad  ``shot\_position": ``Camera position(Options: front\_view/back\_view/side\_view/over\_the\_shoulder/ overhead\_view/point\_of\_view/aerial\_view/overlooking\_view/other)", 
    
\qquad  ``camera\_motion": ``Camera movement description",
    
\qquad  ``environment": ``Video background/environment description",
    
\qquad  ``lighting": ``Lighting information in the video"
    
\}

\end{tcolorbox}

Following the generation of structural captions via the SkyCaptioner-V1 model, we design a caption fusion pipeline to get final captions for text-to-video (T2V) and image-to-video (I2V) model training. Our pipeline utilizes the Qwen2.5-32B-Instruct model to intelligently combine structured caption fields, producing either dense or sparse final captions depending on application requirements.

\begin{tcolorbox}[
    breakable, 
    colback=white, 
    colframe=black!75, 
    title=System Prompt for T2V Prompt Fusion , 
    fonttitle=\bfseries,
    width=\textwidth, 
  ]
You are an expert in video captioning. You are given a structured video caption and you need to compose it to be more natural and fluent in English.

\#\# \textit{Structured Input}

\{structured\_caption\}

\#\# \textit{Notes}

- According to the action field information, change its name field to the subject pronoun in the action.

- If there has an empty field, just ignore it and do not mention it in the output.

- Do not make any semantic changes to the original fields. Please be sure to follow the original meaning.

\#\# \textit{Output Principles and Orders}

- First, declare the shot type, shot angle, shot position if these field are not empty.

- Second, eliminate information in the action field that is not related to the timing action, such as background or environment information.

- Third, describe each subject with its pure action, appearance, expression, position if these fields exist.

- Finally, declare the environment, lighting and camera motion if fields are not empty.

\#\# \textit{Output}

Please directly output the final composed caption without any additional information.

\end{tcolorbox}

\begin{tcolorbox}[
    breakable, 
    colback=white, 
    colframe=black!75, 
    title=System Prompt for I2V Prompt Fusion , 
    fonttitle=\bfseries,
    width=\textwidth, 
  ]
You are an expert in video captioning. You are given a structured video caption and you need to compose it to be more natural and fluent in English.

\#\# \textit{Structured Input}

\{structured\_caption\}

\#\# \textit{Notes}

- If there has an empty field, just ignore it and do not mention it in the output.

- Do not make any semantic changes to the original fields. Please be sure to follow the original meaning.

- If the action field is not empty, eliminate the irrelevant information in the action field that is not related to the timing action(such as wearings, background and environment information) to make a pure action field.

\#\# \textit{Output Principles and Orders}

- First, eliminate the static information in the action field that is not related to the timing action, such as background or environment information.

- Second, describe each subject with its pure action and expression if these fields exist.

- Finally, add camera motion field to the final composed caption if the camera motion field is not empty.

\#\# \textit{Output}

Please directly output the final composed caption without any additional information.

\end{tcolorbox}


\newpage
\section{Video Motion Quality Scoring Criteria For Human Annotation}
\label{app:video_motion_quality_assessment}

\begin{table}[h]
\centering
\small
\renewcommand{\arraystretch}{1.2}
\setlength{\tabcolsep}{12pt}
\begin{tabular}{>{\raggedright\arraybackslash\bfseries}p{4cm}>{\raggedright\arraybackslash}p{11cm}}
\toprule
\multicolumn{2}{l}{\large\bfseries Video Motion Quality Assessment Criteria} \\
\midrule

\textbf{Insufficient Motion Amplitude} \newline (1 point/instance) & 
Subject's motion range significantly less than reasonable real-world range\\[0.5ex]
& Rigid limb movements (insufficient arm swing)\\[0.5ex]
& Facial expressions lack detail (insufficient smile uplift)\\[0.5ex]
& Object trajectories too gentle (vehicles moving too slowly)\\[0.5ex]
\midrule

\textbf{Excessive Motion Amplitude} \newline (2 points/instance) & 
Motion speed/range exceeds realistic physical laws or human behavior\\[0.5ex]
& Abnormal object speeds (car instantly accelerating to extreme speed)\\[0.5ex]
& Exaggerated limb movements (arm bending beyond physiological limits)\\[0.5ex]
\midrule

\textbf{Subject Distortion} \newline (3 points/instance) & 
Unnatural distortion of subject shape/structure during movement\\[0.5ex]
& Body parts stretching during walking (waist becoming thin)\\[0.5ex]
& Legs separating from torso while running\\[0.5ex]
& Surface wrinkles or tears when objects rotate\\[0.5ex]
& Facial muscles causing misaligned features\\[0.5ex]
\midrule

\textbf{Local Detail Distortion} \newline (1 point/instance) & 
Local areas appear blurry, incorrect, or missing\\[0.5ex]
& Hair breaks or color blocks disappear during movement\\[0.5ex]
& Blurry object surface textures (unclear fabric wrinkles)\\[0.5ex]
& Unrecognizable text or identifiers in background\\[0.5ex]
& Vehicle wheels remaining stationary while driving\\[0.5ex]
\midrule

\textbf{Basic Physics Violations} \newline (3 points/instance) & 
Content violates basic laws of physics\\[0.5ex]
& No physical feedback after collisions (ball passing through wall)\\[0.5ex]
& Fluid movements inconsistent with fluid dynamics\\[0.5ex]
\midrule

\textbf{Interaction Violations} \newline (2 points/instance) & 
Subject interactions with environment/other subjects defy reality\\[0.5ex]
& Unreasonable object collisions/penetrations (person passing through closed door)\\[0.5ex]
& Objects colliding without reaction (balls not bouncing after collision)\\[0.5ex]
& Hand-object interaction misalignment (hand not touching cup handle when holding)\\[0.5ex]
\midrule

\textbf{Unnatural/Monotonous Movement} \newline (1 point/instance) & 
Movements lack fluidity or diversity\\[0.5ex]
& Single actions repeated back and forth\\[0.5ex]
& Movements not following conventional paths\\[0.5ex]
& Objects moving in single trajectories (straight lines only)\\[0.5ex]
& Abrupt movement transitions (sudden acceleration/deceleration)\\[0.5ex]
\bottomrule
\end{tabular}
\caption{\textbf{Video Motion Quality Assessment Criteria.} This scoring system identifies and quantifies motion-related issues in generated videos, with point values assigned based on severity and type of motion problem.}
\label{tab:motion_quality_assessment}
\end{table}

\end{document}